\documentclass[runningheads]{llncs}

\usepackage{eccv}

\usepackage{eccvabbrv}

\usepackage{graphicx}
\usepackage{booktabs}

\usepackage{epsfig}
\usepackage{amsmath}
\usepackage{amssymb}
\usepackage[width=122mm,left=12mm,paperwidth=146mm,height=193mm,top=12mm,paperheight=217mm]{geometry}
\usepackage{subcaption}

\usepackage{makecell}
\usepackage{epsfig}
\usepackage{algorithm}

\usepackage{threeparttable}
\usepackage{booktabs}
\usepackage{multicol}
\usepackage{multirow} 
\usepackage{adjustbox}
\usepackage{graphicx} 
\usepackage{colortbl}
\usepackage{listings}

\definecolor{lightyellow}{RGB}{255,242,204}
\definecolor{lightorange}{RGB}{251,229,214}
\definecolor{lightgreen}{RGB}{226,240,217}
\definecolor{lightblue}{RGB}{222,235,247}
\definecolor{lightgray}{RGB}{209,201,206}
\definecolor{deepgray}{RGB}{178,164,173}
\definecolor{deepblue}{RGB}{112,168,218}

\newfloat{lstfloat}{htbp}{lop}
\floatname{lstfloat}{Algorithm}

\lstdefinestyle{myverbatim}{
    basicstyle=\ttfamily\footnotesize,
    backgroundcolor=\color{white},
    breaklines=true,
    breakatwhitespace=true
}

\usepackage[accsupp]{axessibility}  

\usepackage{hyperref}

\usepackage{orcidlink}
\newcommand{\minisection}[1]{\vspace{2mm}\noindent{\textbf{#1}}}
\usepackage{afterpage}
\begin{document}
\title{FlexEdit: Flexible and Controllable Diffusion-based Object-centric Image Editing} 
\titlerunning{FlexEdit: Flexible and Controllable Object-centric Image Editing}
\author{Trong-Tung Nguyen\inst{1} \and 
Duc-Anh Nguyen\inst{1} \and
Anh Tran\inst{1} \and
Cuong Pham\inst{1,2}}

\authorrunning{Trong-Tung Nguyen et al.}
\institute{\hbox{VinAI Research, Vietnam \and
Posts \& Telecom. Institute of Tech., Vietnam}
\email{\{v.tungnt132,v.anhnd72,v.anhtt152,v.cuongpv11\}@vinai.io}}
\maketitle
\vspace{-0.6cm}
\begin{figure}[h]
    \centering
    \includegraphics[width=\textwidth]{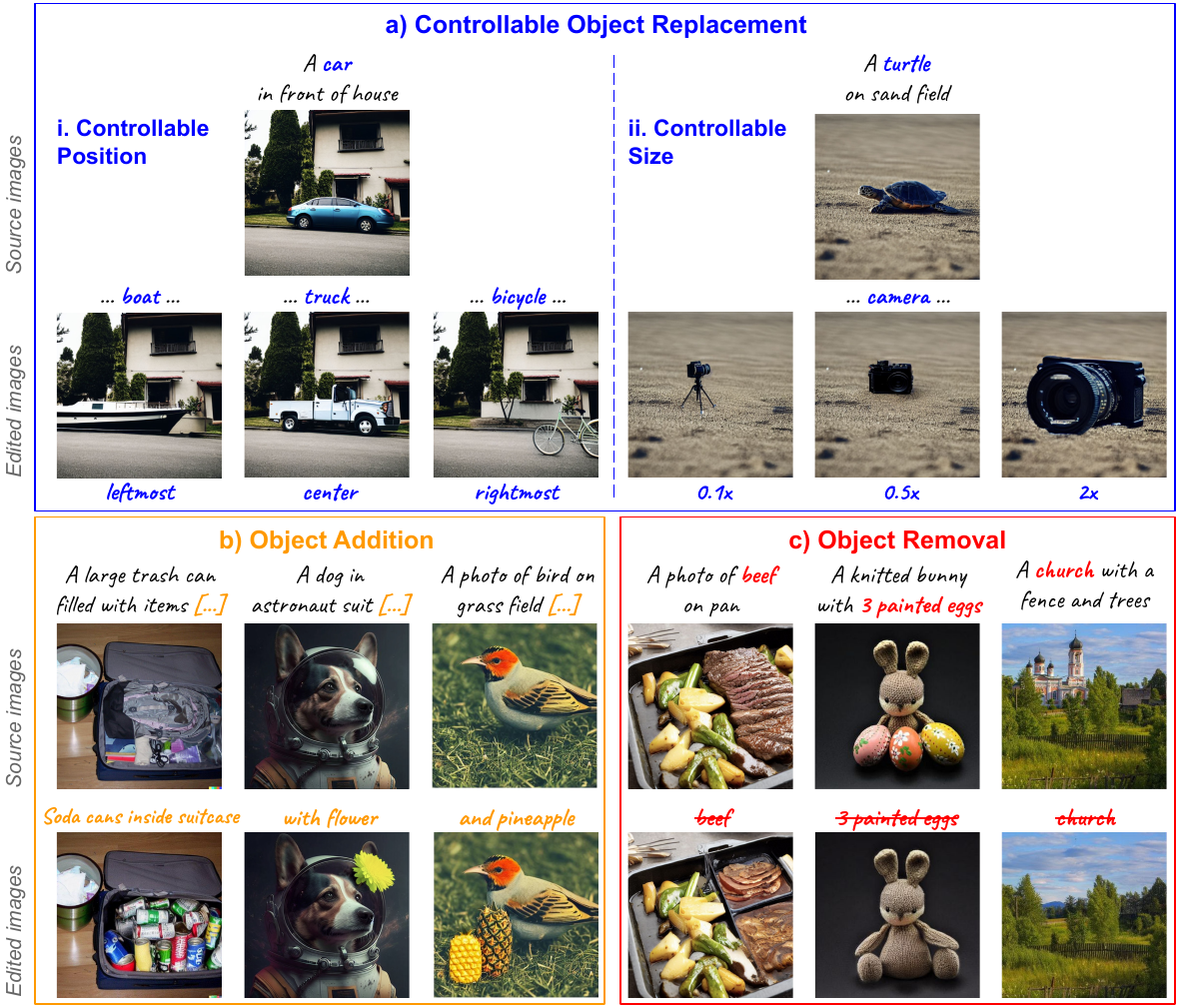}
    \caption{
        Our framework could achieve robust and flexible control over several text-guided object-centric editing scenarios, including a) replacing objects \textbf{with controllable size and position}, b) adding new objects in a natural way \textbf{without additional mask input}, and c) removing objects without compromising the quality of the original image.
    }
    \label{fig:teaser}
\end{figure}
\vspace{-0.9cm}
\begin{abstract}
    Our work addresses limitations seen in previous approaches for object-centric editing problems, such as unrealistic results due to shape discrepancies and limited control in object replacement or insertion. To this end, we introduce FlexEdit, a flexible and controllable editing framework for objects where we iteratively adjust latents at each denoising step using our FlexEdit block. Initially, we optimize latents at test time to align with specified object constraints. Then, our framework employs an adaptive mask, automatically extracted during denoising, to protect the background while seamlessly blending new content into the target image. We demonstrate the versatility of FlexEdit in various object editing tasks and curate an evaluation test suite with samples from both real and synthetic images, along with novel evaluation metrics designed for object-centric editing. We conduct extensive experiments on different editing scenarios, demonstrating the superiority of our editing framework over recent advanced text-guided image editing methods. Our project page is published at \url{https://flex-edit.github.io/}.
\end{abstract}
\section{Introduction}
\label{sec:intro}
Text-to-image generation, fueled by recent advancements in large-scale generative diffusion models trained on extensive image-text pairs datasets, has recently become a focal point of research. Models like Imagen \cite{saharia2022photorealistic}, DALLE-2 \cite{ramesh2022hierarchical}, and Stable Diffusion \cite{rombach2022highresolution} excel not only in conditional image generation but also offer rich priors beneficial to visual content creation tasks. Among these tasks, text-guided image editing stands out, demanding edited images to maintain fidelity to the source while incorporating desired edits guided by text prompts.

Recent approaches leverage powerful generative models as priors for image editing. Several methods \cite{hertz2022prompttoprompt,mokady2022nulltext,cao2023masactrl,tumanyan2022plugandplay} employ attention-hijacking mechanisms, alternating between source and target edit content during denoising to achieve extensive control over edits while preserving source content. Others \cite{kawar2023imagic,bartal2022text2live,zhang2022sine,valevski2023unitune} maintain content preservation via fine-tuning before a denoising process guided by text prompts. In contrast, DiffEdit\cite{couairon2022diffedit} and WatchUrStep\cite{mirzaei2023watch} integrate a coarse editing mask estimated by the discrepancy between noise maps conditioned on two different text prompts. However, these methods have limitations, particularly in object-centric editing scenarios. For instance, when replacing objects, the edited objects may not align with the expected class due to differences in size or shape. This lack of consistency hinders real-world editing applications, especially when assumptions about properties like size or position cannot be made. Moreover, these methods show inconsistent results in other object-centric editing tasks like addition and removal, which are essential in real-world settings.

To overcome these limitations, we present FlexEdit, a diffusion-based framework tailored for object-centric image editing. Built on the Stable Diffusion model, FlexEdit integrates advanced components for flexible and precise object editing across diverse scenarios. At each denoising step, we combine two essential elements: latent optimization and blending with an adaptive binary object mask. Initially, we refine noisy latent codes with loss functions incorporating multiple object constraints. Then, we utilize an automatically generated adaptive object mask to blend edited visual contents with background information based on editing specifications. These processes are iterated ensure that noisy latents maintain editing semantics while preserving fidelity to the source image.

\begin{figure}[t]
    \centering
    \includegraphics[width=\textwidth]{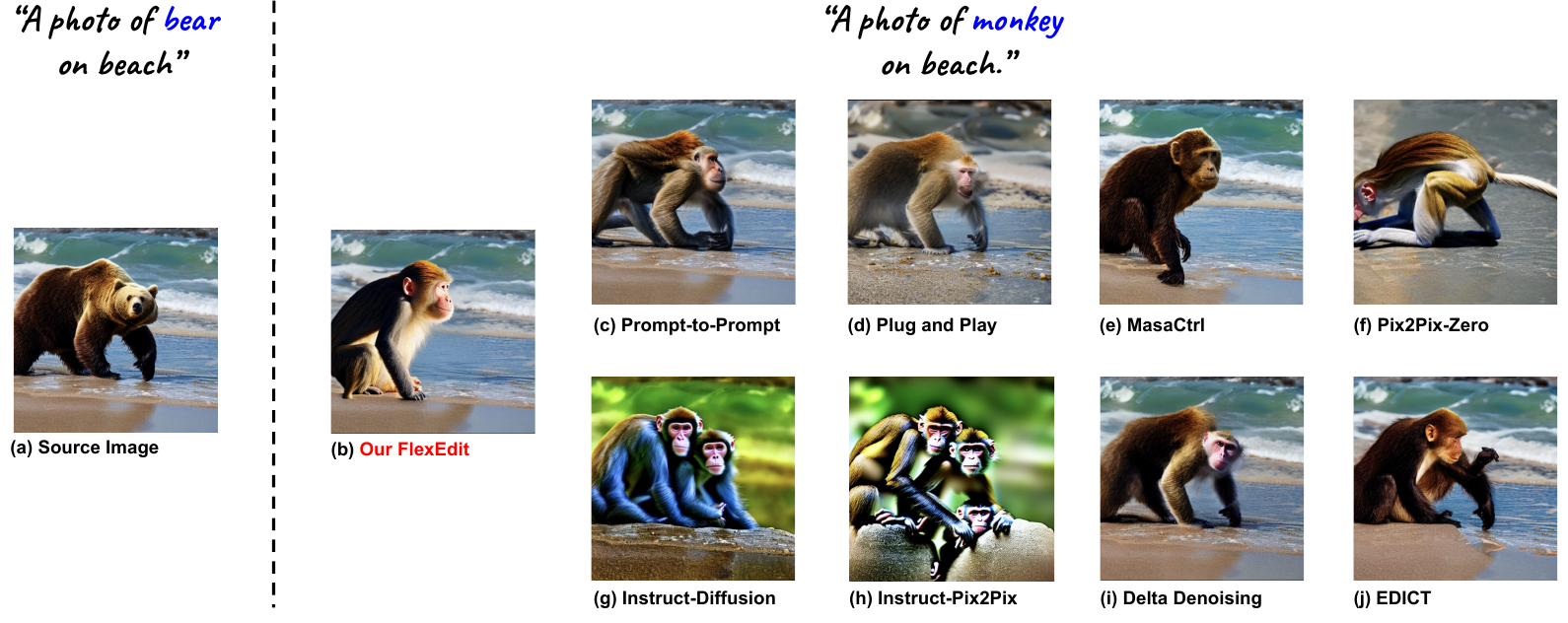}
    \vspace{-2mm}
    \caption{
        We show an editing scenario when edited object \textcolor{blue}{monkey} and source object \textcolor{blue}{bear} are distinct in shape. Our FlexEdit could achieve flexible shape transformation editing while preserving high fidelity to the source image's background information.
    }
    \vspace{-4mm}
    \label{fig:previous_limitation}
\end{figure}

To assess our editing framework's effectiveness, we utilize two established benchmarks: MagicBrush \cite{zhang2023magicbrush} and PieBench \cite{ju2023direct}. From these, we create two subsets, MagicO and PieBenchO, comprising samples relevant to object-centric editing scenarios like addition, replacement, and removal. Additionally, we introduce a new test suite, SynO (Synthetic Object-centric editing), tailored for object-centric synthesized image editing, covering diversified samples. Recognizing the needs for suitable evaluation metrics, we also propose a new metric set specifically designed for object-centric editing. Using these benchmarks, we conduct extensive experiments on recent state-of-the-art editing algorithms, demonstrating our method's consistency in achieving a balance between background preservation and editing semantics. A user study further confirms user preference for our editing results compared to others. Our contributions are threefold:
\begin{itemize}
    \item We propose a new editing framework for object-centric image editing tasks.
    \item We introduce a novel test suite, including test samples along with new evaluation metrics for our editing problem.
    \item We provide an extensive evaluation on different benchmarks and various state-of-the-art methods to showcase the versatility of our editing framework in various flexible and customizable object editing applications.
\end{itemize}



\section{Related Works}
\label{related_works}

\subsection{Text-guided Image Editing with Diffusion Model}
Text-guided image editing is a challenging task that aims to edit a given source image to make it adhere to a textual input while preserve unrelated background content. Following the success of diffusion-based text-to-image generation models, particularly open-source models like Stable Diffusion (SD), recent works aim to exploit them for text-guided image editing tasks. Most techniques harness SD's powerful attention mechanism, which builds up its strong connection between text and generated images. For instance, P2P\cite{hertz2022prompttoprompt} leverages a strong connection between image and text observed in the attention layers of the denoising UNet model and flexibly control the attention map to facilitate text-guided image editing. Plug-and-Play \cite{tumanyan2022plugandplay} achieves fine-grained control to translate image by adjusting intermediate features and self-attention layers. This aims to align translated content with specified text prompts while preserving layout guidance. MasaCtrl \cite{cao2023masactrl} proposes to convert existing self-attention layers into mutual self-attention to query correlated local contents and textures from source images that could achieve simultaneous image generation and non-rigid image editing. 

When editing real images, a critical pre-processing step is inversion, i.e., finding the suitable noise latent input that can reconstruct the image through the denoising process. Several inversion methods have been proposed. SDEdit \cite{meng2022sdedit} adds random Gaussian noise to source images as input but struggles with reconstruction quality and accurate local editing. DDIM Inversion \cite{song2022denoising} provides a rough approximation of the source image but deviates significantly from the original trajectory at large guidance scales, compromising source image preservation. Null-text Inversion \cite{mokady2022nulltext} iteratively optimizes the ``null'' text in the classifier-free guidance procedure aiming to minimize the L2 distance between pivotal and target trajectories at each denoising step but is computationally expensive. Direct-Inversion \cite{ju2023direct} disentangles the diffusion path into source and target branches, avoiding optimization procedures and achieving practical running times.

These approaches, however, often lack of flexible and controllable editing abilities, particularly for objects. Hence, we propose a novel editing framework focused on a wide range of flexible and controllable object-specific editing scenarios.
\subsection{Controllable Image Synthesis}
Recent image synthesis approaches leverage pre-trained diffusion models to enhance image fidelity or provide additional control over generated content. For instance, Attend-Excite \cite{chefer2023attendandexcite} solves the issue of catastrophic neglect occurring when generating more than two objects from the input prompt. They propose to optimize the latent representation at each denoising step to enhance the visual appearance of object fidelity given in input text prompt. Similarly, BoxDiff \cite{xie2023boxdiff} allows users to specify the expected location where objects should be rendered and then performs optimizing the latent at each step using constraints designed for object boxes. Additionally, Self-Guidance \cite{epstein2023diffusion} introduces guidance functions to steer the sampling process based on object properties such as size, shape, and position. However, these advancements have not been fully applied to image editing. Our aim is to incorporate it into image editing, enabling users to perform a variety of controllable object-centric editing.

\section{Background}\label{sec:background}
\subsection{Stable Diffusion Model}
Diffusion model (DM) is a powerful generative framework that aims to generate high-quality images via denoising process. However, it shows limitations in long running times due to large computational space. Latent Diffusion Model (LDM)\cite{rombach2022highresolution} was introduced to address the issue. It encodes the original input and brings the denoising process into latent space instead of pixel space, hence improving time efficiency. Stable Diffusion (SD) is an implementation built upon LDM to enable high-quality text-to-image generation. With SD, an input image $x$ is first encoded into latent code $z=\mathcal{E}(x)$ with encoder $\mathcal{E}$ and decoded with a decoder $\mathcal{D}$ to reconstruct image  $\hat{x}=\mathcal{D}(\mathcal{E}(x))$. SD also uses an additional text prompt input $y$, which is often preprocessed and then encoded by a text-encoder $\tau_{\theta}(y)$ of pre-trained CLIP\cite{radford2021learning}. 
Its core network component is a denoising UNet trained to denoise any noisy latents $z_t$ at time step $t \in \{1, 2, ..., T\}$, with $T$ is the number of time steps, conditioned on text embedding $\tau_{\theta}(y)$ to get a cleaner latent at the previous time step $t - 1$. During inference, we can sample $z_T$ from $\mathcal{N}(0,1)$ and then start the denoising process initializing with $z_T$ until we reach $z_0$, then feed it into the decoder $\mathcal{D}$ to synthesize an image $\hat{x}$.

\subsection{Cross-Attention and Self-Attention Layers in SD}
The text encoder $\tau_{\theta}$ discussed above projects text input $y$ of length $N$ to an intermediate representation $\tau_{\theta}(y) \in \mathbb{R}^{N \times M}$ with $M$ is the feature size. It is then mapped to cross-attention map $l^{th}$ at time step $t$. This cross-attention map demonstrates a strong connection between visual and textual information. Besides cross-attention maps, UNet is also equipped with self-attention maps which are commonly used in practice to capture the self-information of the input being associated with itself. In general, the cross-attention and self-attention map at layer $l^{th}$ and time step $t$ could be obtained via:
\begin{equation}
\mathcal{A}^C_{l,t}=\operatorname{SM}\left(\frac{Q_{z} K_{e}^{\top}}{\sqrt{d_l}}\right) \in[0,1]^{HW \times N},\;
\mathcal{A}^S_{l,t}=\operatorname{SM}\left(\frac{Q_{z} K_{z}^{\top}}{\sqrt{d_l}}\right) \in[0,1]^{HW \times HW},\!\label{eq:self_map}
\end{equation}
where $\operatorname{SM}(\cdot)$ is the softmax function, $d_l$ is the number of features at layer $l$. Here, the query $Q_z$ and key $K_z$ are different projections from the flattened intermediate representation of $z_t$ while the key $K_e$ is projected from the text embedding  $\tau_{\theta}(y)$.


\section{Approach}
In this section, we first outline our diffusion-based editing framework (\cref{overview}). Then, we discuss a core technique of extracting fine-grained object masks from attention maps during the diffusion process (\cref{sec:dynamic_mask}) before delving into the algorithm details. Our method manipulates the latent representation for editing at each denoising time step using a FlexEdit block that consists of two components: a latent optimization process to acquire the editing semantics (\cref{sec:optimization}) and a latent blending procedure to preserve background content (\cref{sec:blended_latent}). These components are iteratively executed (\cref{sec:iterative}). Our pipeline is shown in \cref{fig:diagram}.
\subsection{Overview of Editing Framework}
\label{overview}
\begin{figure}[t]
    \centering
    \includegraphics[width=\textwidth]{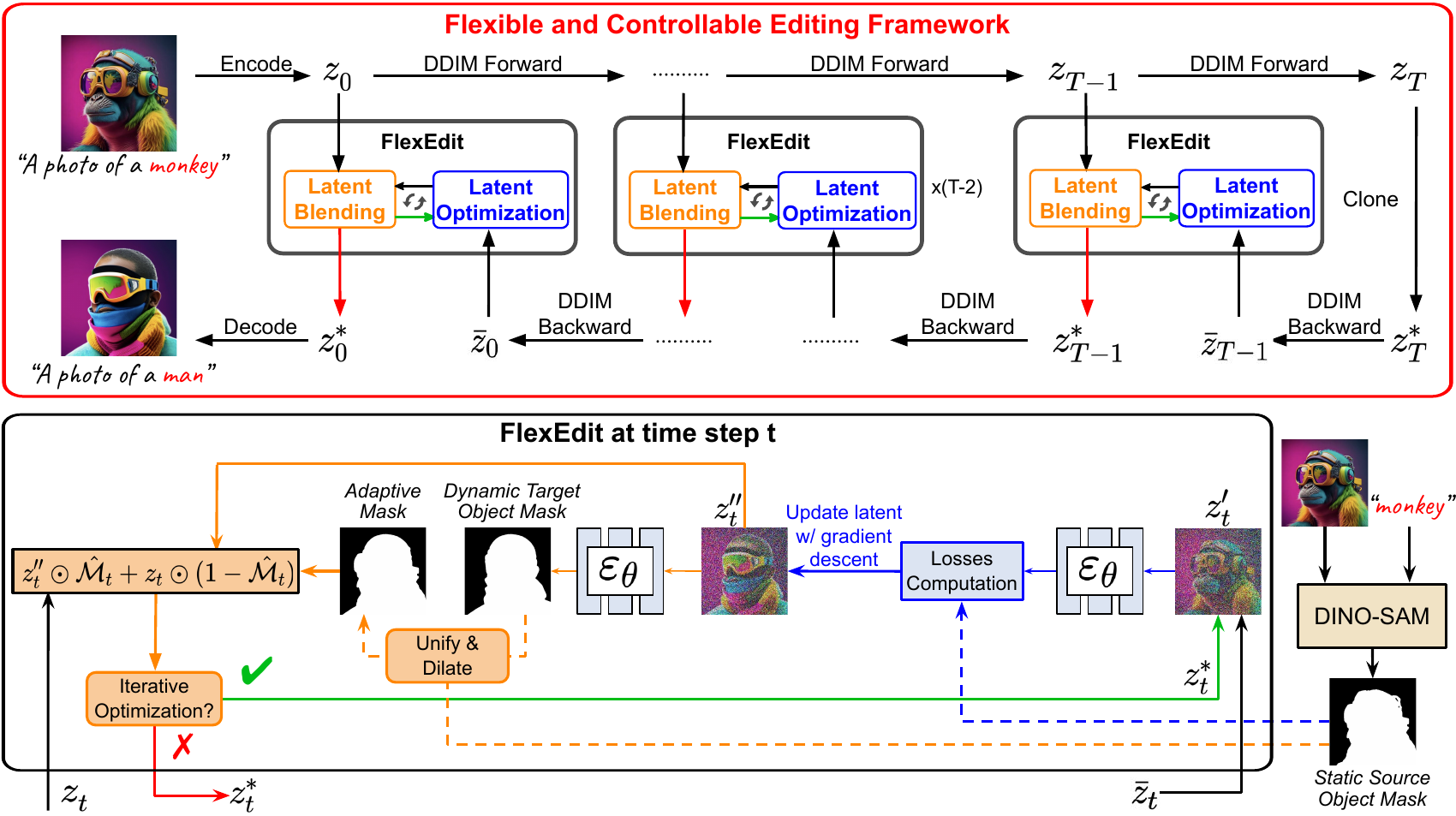}
    \vspace{-2mm}
    \caption{
        \textbf{Overview of FlexEdit framework.} Given an input image $I$, we first bring it to the intermediate source latents through an inversion process. Subsequently, the denoising process starts from ${z}^*_T$ cloned from $z_T$ after the inversion process and progresses toward $z_0^*$, which is then decoded to get the edited image $I^*$. At each denoising step, our FlexEdit block manipulates the noisy latent code through two main submodules: latent optimization (shown in \textcolor{blue}{blue}), and latent blending (shown in \textcolor{orange}{orange}). This is to achieve editing semantics as well as to maintain high fidelity to the source image. If the iterative process (shown in \textcolor{green}{green}) is not executed, our FlexEdit would return $z_t^*$.}
    \vspace{-3mm}
    \label{fig:diagram}
\end{figure}
Our framework achieves image editing via two main stages: a forwarding stage that collects noisy latents corresponding to the source image at every time step and an editing stage that gradually manipulates and denoises noisy latents to generate the edited image. It is noteworthy that our editing framework is applicable to both real and synthesized image editing. The distinction lies in the first stage. With real image editing, we employ a straightforward inversion process, specifically DDIM Inversion\cite{song2022denoising}, to gradually add noise to the initial source latent $z_0$ encoded from source image $I$. As shown in \cref{fig:diagram}, the inversion process brings $z_0$ through intermediate source latents $z_t$ at different noise levels for time step $t$ until a predetermined ending time step $T$, resulting in $z_T$. With synthesized image editing, obtaining $z_t$ is straightforward as we can record them during the denoising process when generating the synthesized image $I$.
These intermediate source latents $\{z_t\}_{t=0}^T$ play a crucial role in our editing framework, as will be discussed in \cref{sec:blended_latent}.
Starting from $z_T^*$ cloned from $z_T$,  the editing process gradually denoises noisy latents to bring $z_{t+1}^*$ into $z_t^*$ where our FlexEdit block is integrated at each time step. Finally, the resulting clean latent $z_0^*$ is obtained and then decoded to generate an edited image $I^*$. 

\subsection{Dynamic Object Binary Mask from Attention Map}
\label{sec:dynamic_mask}
Based on \cref{eq:self_map}, we could extract the cross-attention map $\mathcal{A}^{C}_{j, l, t}$ for object token $j^{th}$ and self-attention map $\mathcal{A}^{S}_{l, t}$ at each time step $t$ where we choose the resolution of $16 \times 16$ and $32 \times 32$ respectively. These resolutions have been shown to contain the most semantic information in \cite{hertz2022prompttoprompt}. We then aggregate across all layers to obtain the average cross-attention and self-attention maps via:
\begin{align}
    \mathcal{A}^{C}_{j, t} = \frac{1}{L} \sum_{l=1}^{L} \mathcal{A}^{C}_{j, l, t},\quad \quad \quad \quad
    \mathcal{A}^{S}_{t} = \frac{1}{L} \sum_{l=1}^{L} \mathcal{A}^{S}_{l, t}.
    \label{eq:average_attention}
\end{align}

\begin{figure}[t]
    \centering
    \includegraphics[width=\textwidth]{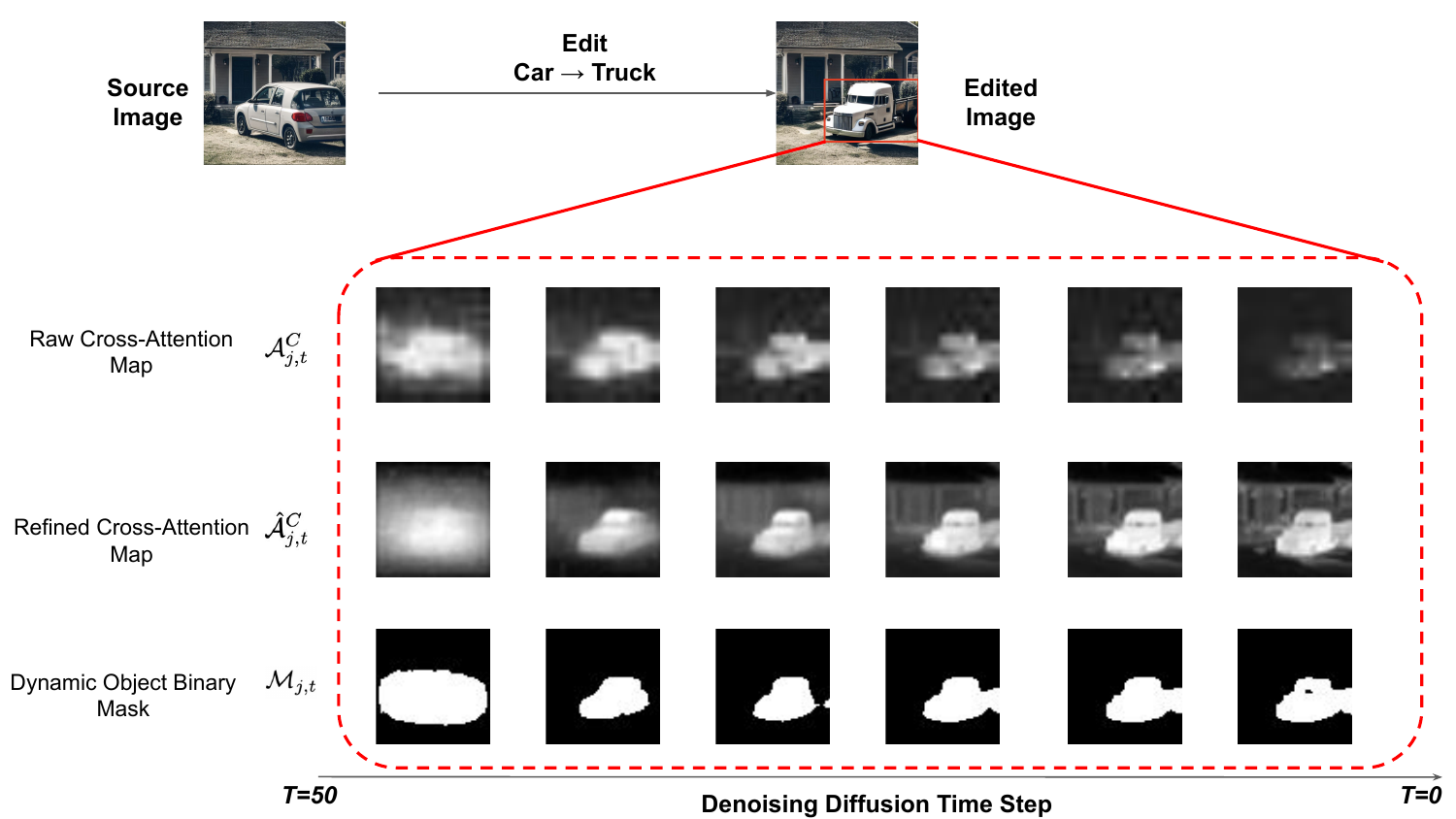}
    \vspace{-2mm}
    \caption{Visualization of different versions of cross-attention maps and dynamic binary masks for edited object, i.e. truck during the denoising diffusion process.}
    \label{fig:dynamic_mask_visualization}
    \vspace{-5mm}
\end{figure}

As shown in the first row of \cref{fig:dynamic_mask_visualization}, $\mathcal{A}^{C}_{j, t}$ are coarse-grained and fail to indicate exact segmentation for the object \textit{truck} at every single time step $t$.
Inspired from \cite{nguyen2023dataset}, we alleviate this issue by combining $\mathcal{A}^{C}_{j, t}$ with $\mathcal{A}^{S}_{t}$ to enhance its granularity. In this way, $\mathcal{A}^{C}_{j, t}$ could benefit from the self-information being propagated to associated similar object regions contained in $\mathcal{A}^{S}_{t}$. Formally, we obtain the refined cross-attention map $\hat{\mathcal{A}}_{j, t}^C$ by taking exponential of the average self-attention map $\mathcal{A}^{S}_{t}$ to the power of $\tau$ before multiplying it with $\mathcal{A}^{C}_{j, t}$ via
\begin{equation}
    \hat{\mathcal{A}}_{j, t}^C = (\mathcal{A}_{t}^S)^{\tau} \cdot \mathcal{A}_{j, t}^C, \qquad \qquad \mathcal{A}_{j, t}^C \in [0, 1]^{16\times16}.
    \label{eq:exponentiation}
\end{equation}

As demonstrated in the second row of \cref{fig:dynamic_mask_visualization}, $\hat{\mathcal{A}}_{j, t}^C$ provides more precise activation of object \textit{truck}. Finally, we use a threshold cut-off $\beta$ to convert $\hat{\mathcal{A}}^{C}_{j, t}$ into a binary mask $\mathcal{M}_{j, t}$ as shown in third row of \cref{fig:dynamic_mask_visualization}, which could be used to separate the foreground object $j^{th}$ and the background region. We utilize $\mathcal{M}_{j, t}$ across various stages in our FlexEdit block, as detailed in the following sections.

\subsection{Latent Optimization with Object Constraints}
\label{sec:optimization}

We first thoroughly analyze critical attributes of editing object, aiming to achieve high-quality visual appearance and adherence to user-defined object constraints. Whether replacing, adding, or removing objects specified by the editing text prompt $P^*$, each setting presents unique challenges requiring tailored manipulation of noisy latents for realistic results. In the following subsection, we discuss how to address these challenges by introducing various object constraints.



\minisection{Controllable Object Replacement.}
Without imposing constraints on the edited object's properties regarding size and position, the edited objects may exhibit high randomness and may not align with the user's editing intentions. To this end, we utilize attention-based estimation methods from \cite{epstein2023diffusion} to approximate the size and position of the edited object. Subsequently, we obtain the expected position or size of the edited object from the user and aim to minimize the discrepancy between these two quantities.
Given the target object specified by the $j^{th}$ textual token, we obtained its dynamic object mask $\mathcal{M}_{j, t}$ as described in \cref{sec:dynamic_mask}. We then calculate its centroid and size as follows:
\begin{equation}
    \texttt{centroid}_{j, t} =\frac{1}{{\sum_{h,w}^{}\mathcal{M}_{j,t,h,w}}} 
    \begin{bmatrix}
        \sum_{h,w} w \cdot \mathcal{M}_{j,t,h,w} \\ 
        \sum_{h,w} h \cdot \mathcal{M}_{j,t,h,w}  
    \end{bmatrix},
    \label{eqn:position}
\end{equation}

\begin{equation}
    \texttt{size}_{j,t} = \frac{1}{HW}\sum_{h,w}^{}\mathcal{M}_{j,t,h,w},
    \qquad \qquad \mathcal{M}_{j,t} \in [0, 1]^{H\times W},
    \label{eqn:size}
\end{equation}
where $\mathcal{M}_{j,t,h,w}$ at coordinates $(h,w)$ denotes activation value of $\mathcal{M}_{j, t}$,
which is resized to match the resolution of the edited image, i.e., $H\times W$. The user could then explicitly provide their expected centroid and size of the edited object, which are denoted as $\texttt{centroid}_t^*$ and $\texttt{size}_t^*$ respectively. To control the position and size of the edited object, we minimize the L2 distance between expected and estimated properties obtained in \cref{eqn:size} and \cref{eqn:position}
via
\begin{equation}
    {\mathcal{L}_{pos}}=\|\texttt{centroid}_{j,t} - \texttt{centroid}_t^*\|_2^2, \qquad {\mathcal{L}_{size}}=\|\texttt{size}_{j,t} - \texttt{size}_t^*\|_2^2.
    \label{eqn:loss_property}
\end{equation}

\minisection{Attention Separation in Mask-free Object Addition.}
The challenge when inserting objects in a mask-free setting is the attention overlapping problem, as discussed in \cite{li2023divide, agarwal2023astar}, which occurs when model is confused in allocating similar regions to different objects. We mitigate such issue with a separation constraint to delineate regions for each object.
Consider a single existing object $i^{th}$, we first extract its binary mask $\mathcal{S}_i$ using an off-the-self segmentation module DINO-SAM \cite{ren2024grounded}. For edited object $j^{th}$, we utilize our dynamic mask $\mathcal{M}_{j,t}$ extracted in \cref{sec:dynamic_mask}. and then aim to separate $\mathcal{M}_{j,t}$ from $\mathcal{S}_i$ via a loss that minimizes their similarity. Intuitively, minimizing their overlapping area is an option but this does not always work since these two regions can be fragmented and interleaved. Hence, the overlapping area is zero, while the object masks are not fully separated. Instead, we empirically found cosine similarity to be a reliable metric for the separation loss. Formally, we flatten $\mathcal{M}_{j,t}$ and $\mathcal{S}_i$ into two vectors denoting as $f_{j,t}$ and $g_{i}$, respectively; and then compute their cosine similarity via
\begin{equation}
    {\mathcal{L}_{sep}}=\frac{\sum_{k=1}^{H \times W}f_{j,t}^{k}.g_{i}^{k}}{\left \| f_{j,t} \right \|^2_2.\left \| g_{i} \right \|^2_2}.
    \label{eqn:loss_separation}
\end{equation}
A high loss value $\mathcal{L}_{sep}$ suggests that values within the vectors are activated in the same region, while a low loss value indicates separate regions. Minimizing this loss ensures that the added object does not interfere with existing objects.

\minisection{Latent Optimization via Object Constraints.}
We observe that each editing scenario demands different object constraints, which we aim to utilize to perform latent optimization. For object replacement, we offer flexible property control via two controllable losses shown in \cref{eqn:loss_property}. In object addition, we address attention overlapping using separation loss discussed in \cref{eqn:loss_separation}. For object removal, no optimization is needed as no target objects are generated; thus, no constraints are imposed. We use ${\mathcal{L}}_{optim}$ as the representative loss for these editing scenarios.
At each time step, we denote $z'_t$ as the noisy latent before the optimization step. We pass $z'_t$ through the denoising UNet model $\epsilon_{\theta}(.)$ to generate attention maps which are used to compute loss functions discussed as above, resulting in a representative loss ${\mathcal{L}}_{optim}$. We then update $z'_t$ via applying gradient descent with a scaling factor $\alpha_t$ into $z^{'}_t$ via
\begin{equation}
    z^{''}_t \leftarrow z^{'}_t - \alpha_t.\nabla_{z_{t}^{'}}{\mathcal{L}}_{optim}.
\label{eq:optimization}
\end{equation}

\subsection{Latent Blending with Adaptive Binary Mask}
\label{sec:blended_latent}


The resulting noisy latent $z_{t}^{''}$ may lose background information of the source image. Therefore, we address such issue by utilizing source intermediate latents $z_t$ as discussed in \cref{overview}. Specifically, we blend $z^{''}_t$ with the corresponding source latent $z_t$ via our adaptive binary mask $\hat{\mathcal{M}_{t}}$ as follows:
\begin{equation}
    z^{*}_t \leftarrow z^{''}_t\odot \hat{\mathcal{M}_{t}} + z_t\odot (1-\hat{\mathcal{M}_{t}}).
\end{equation}

The adaptive mask $\hat{\mathcal{M}_{t}}$ is constructed via combining both source and target object regions to ensure both flexibility and accurate background preservation. This is built upon our observation in various object-centric image editing scenarios such as (1) transforming a source object into a target one (object replacement), (2) removing a source object (object removal), or (3) adding an extra object without affecting existing ones (object addition). Relying solely on one object region may limit shape transformation or result in incomplete edits. 

Specifically, we reused the source object mask $\mathcal{S}_i$ discussed in \cref{sec:optimization} and the dynamic target object mask $\mathcal{M}_{j,t}$ in \cref{sec:dynamic_mask}. The adaptive object binary mask is constructed by applying dilation operation $f_{dilate}$ on the unified region: 
\begin{equation}
    \hat{\mathcal{M}_{t}} = f_{dilate}(\bigcup_{i\in{O^S}}^{}\mathcal{S}_i \cup \bigcup_{j\in{O^T}}^{}\mathcal{M}_{j,t}),
\end{equation}
where $O^S$ and $O^T$ are the set of all source and target object tokens, respectively. The dilation function 
$f_{dilate}$ is helpful in slightly extending the mask, avoiding visible seams observed in the edited region due to an overly tight mask.


\subsection{Iterative Latent Manipulation with FlexEdit}\label{sec:iterative}

As discussed in \cite{chefer2023attendandexcite}, single-step latent optimization does not guarantee low loss values. Hence, we iterate the processes of latent optimization (\cref{sec:optimization}) and latent blending (\cref{sec:blended_latent}). In detail, we perform such iterative latent manipulation at time step $t_1=1, t_2=10, t_3=15, t_4=20$  and set different criteria for finishing iteration based on the value of the loss terms (\cref{sec:exp_setup}).

\section{Experiments}
\subsection{Experimental Setup}
\label{sec:exp_setup}
\minisection{Evaluation Datasets.}
For real image editing, we used two recent evaluation sets: MagicBrush\cite{zhang2023magicbrush} and PieBench\cite{ju2023direct}. MagicBrush\cite{zhang2023magicbrush} covers various edit instructions and captions across different editing scenarios, while PieBench \cite{ju2023direct} focuses on language-driven evaluation featuring ten editing types. We curated their samples tailored for the object-centric image editing problems, forming two subsets: MagicO (from MagicBrush) and PiebenchO (from PieBench). MagicO includes 254 samples, with 53 for object replacement, 187 for object addition, and 14 for object removal. PieBenchO consists of 217 samples, with 75 for object replacement, 69 for object addition, and 73 for object removal. 

Existing benchmarks often focus on real image inputs, neglecting the effect of inversion inaccuracies when editing. We address such issue by additionally introducing a new test suite called SynO designed specifically for synthetic image editing which covers various object-centric editing scenarios, including replacement, addition, and removal. Each sample includes a synthesized image, original and target prompts, and an equivalent edit instruction. We ensure diversity by incorporating different object transformations, reflecting the variability of objects in real-world settings. In detail, SynO consists of 1079 editing samples for object replacement, 483 for object addition, and 90 for object removal. 

\minisection{Evaluation Metrics.}
\label{sec:evaluation_metrics} 
Background preservation and editing semantics are two crucial aspects of image editing evaluation. While the former ensures fidelity to the source image, the latter measures alignment with the provided text prompt. Existing metrics, as seen in previous works \cite{hertz2023delta,ju2023direct,couairon2022diffedit}, often use LPIPS score to measure background preservation and CLIP score to quantify editing semantics for text-image alignment. They compute LPIPS score on the whole image while measuring CLIP score on the region specified via an annotated mask. However, this may not be suitable for object-centric editing since background preservation should be measured on the background region only. On the other hand, CLIP score should reflect the editing semantic between the editing region of the edited image and editing object tokens. Thus, we propose a novel automatic mask-based evaluation metric tailored to various object-centric editing scenarios.

We define three masks: source object, target object, and background masks. The source object mask locates the source object in the source image, while the target object mask locates the target object in the edited image. In scenarios without a source or target object (object addition or removal), the respective masks are empty. We use DINO-SAM \cite{ren2024grounded} to extract masks for both source object $\mathcal{M}_{src}$ and target object $\mathcal{M}_{tgt}$. The background mask $\mathcal{M}_{bg}$ is derived by complementing the union of the source and target object masks, i.e., $1-(\mathcal{M}_{src} \cup \mathcal{M}_{tgt})$. With these masks, we can evaluate editing results based on two criteria:
\begin{itemize} 
\item Background Preservation: We use LPIPS to measure image difference between the background regions of the source image $I$ and edited one $I^*$: 
\begin{equation}
    \text{LPIPS} = \text{LPIPS}(\mathcal{M}_{bg} \odot I, \mathcal{M}_{bg} \odot I^*).
\end{equation}
\item Editing Semantics: Using CLIP and the object masks, we compute a CLIP-O score that measures the success in producing the target object given the target token $w^* \in P^*$ and a CLIP-NO score that measures the success in removing the source object corresponding to the source object token $w \in P$:
    \begin{align}
            \text{CLIP-O} & = \text{CLIP}(\mathcal{M}_{tgt} \odot I^*, w^*),\\
            \text{CLIP-NO} & = 1-\text{CLIP}(\mathcal{M}_{src} \odot I^*, w).
        \end{align}
Note that CLIP-O is not applicable in the object removal scenario, while CLIP-NO is not applicable in object addition.
\end{itemize}

\minisection{Hyper-parameters.} We set $\tau$ in \cref{eq:exponentiation} as 4, the binary threshold $\beta$ as $0.6$. For the gradient descent factor $\alpha_t$,  we follow \cite{chefer2023attendandexcite} by starting from $\alpha_T=20$ and decaying linearly using a linear scheduling rate to the value of $10$. In the iterative latent manipulation process (\cref{sec:iterative}), we empirically set the stopping criteria as when $\mathcal{L}_{pos}$ and $\mathcal{L}_{size}$ both reach minimum values of $T_1=0.4, T_2=0.2, T_3=0.1$ and $T_4=0.05$ corresponding to four mentioned time steps. For separation loss $\mathcal{L}_{sep}$, we set the corresponding thresholds as $V_1=0.8, V_2=0.5, V_3=0.3$, and $V_4=0.1$. The maximum number of iterations is $20$ to encourage the latent to remain in-distribution if these criteria could not be reached.

\subsection{Experimental Results}
\label{experimental_results}
To show effectiveness, we extensively compare FlexEdit with state-of-the-art methods in both real and synthesized image editing tasks, including caption-based approaches like Plug-and-Play\cite{tumanyan2022plugandplay}, P2P\cite{hertz2022prompttoprompt}, Pix2Pix-Zero\cite{parmar2023zeroshot}, MasaCtrl\cite{cao2023masactrl}, EDICT\cite{wallace2022edict}, DDS\cite{hertz2023delta}, and instruction-based methods like InstructPix2Pix\cite{brooks2023instructpix2pix} and Instruct-Diff\cite{geng2023instructdiffusion}. For methods relying on inversion for real-image editing such as Plug-and-Play\cite{tumanyan2022plugandplay}, P2P\cite{hertz2022prompttoprompt}, Pix2Pix-Zero\cite{parmar2023zeroshot}, MasaCtrl\cite{cao2023masactrl} we report the experimental results based on DDIM Inversion\cite{song2022denoising} for fair comparison. In addition, we also provide full experimental results on other inversion methods such as Direct-Inversion\cite{ju2023direct} and Null-text Inversion\cite{mokady2022nulltext} in the Appendix. 

\begin{figure}[t]
    \centering
    \includegraphics[width=\textwidth]{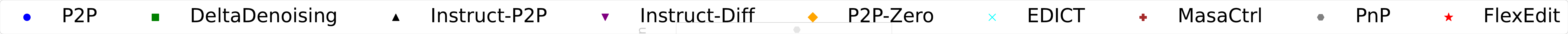}
    \vspace{-1mm}
    \includegraphics[width=\linewidth]{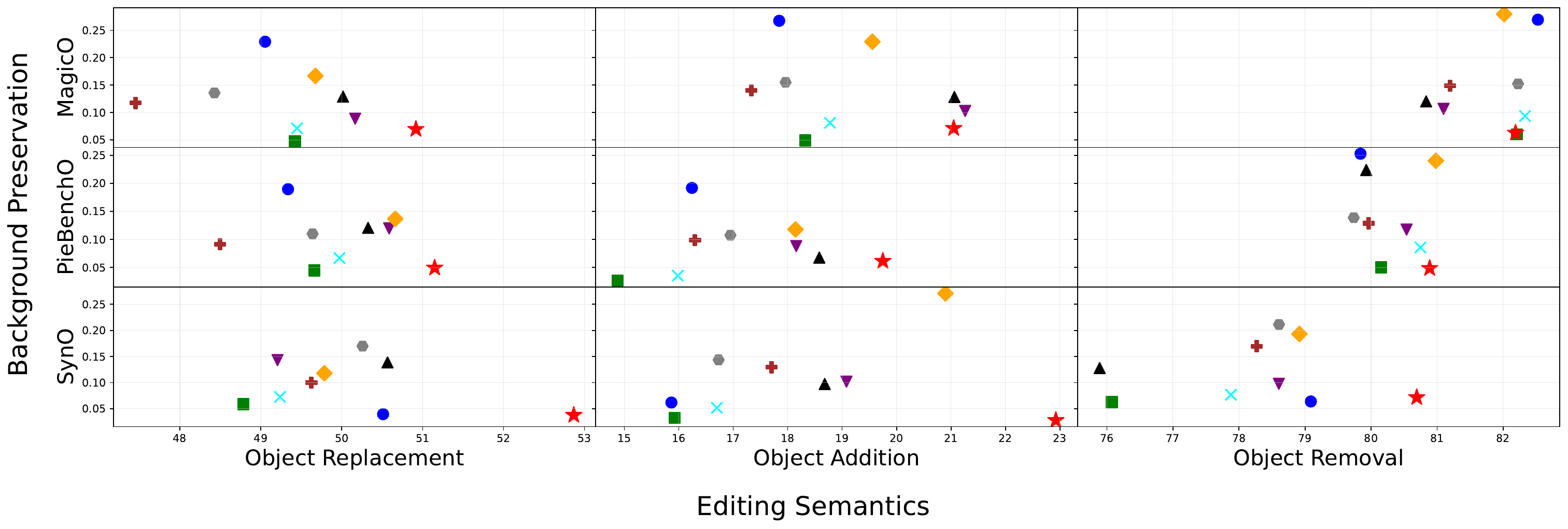}%
\vspace{-3mm}
    \caption{Performance comparison of FlexEdit against existing editing techniques on the SynO, PieBenchO, and MagicO datasets. The method on the bottom right of each subplot provides the best background preservation and editing quality trade-off.}
    \label{fig:scatter_plot}
\vspace{-5mm}
\end{figure}

\minisection{Main Quantitative Results.}
\cref{tab:quantiative_results} show the experimental results of 9 editing methods, including ours, across 3 editing benchmarks and 3 editing tasks. In general, FlexEdit achieves competitive scores in most editing tasks. Some methods like DDS\cite{hertz2023delta}, or EDICT \cite{wallace2022edict} maintain good LPIPS scores as shown in MagicO results but struggle with high editing semantic alignment. Conversely, methods like InstructPix2Pix \cite{brooks2023instructpix2pix} or Instruct-Diff \cite{geng2023instructdiffusion} show promising editing semantics in MagicO but do not achieve reliable LPIPS scores. Our approach demonstrates a superior trade-off between two criteria across various editing scenarios in all three benchmarks. To visually illustrate this, we provide scatter plots in \cref{fig:scatter_plot}, where each entry represents a particular editing scenario. Methods on the bottom right of each plot are considered to balance well between the two criteria. FlexEdit outperforms other methods in achieving a reliable trade-off.

\begin{table}[t]
\caption{
        Quantitative comparison of FlexEdit with other editing methods on three editing benchmarks: SynO, PiebenchO, and MagicO. 
    }
\vspace{-3mm}
\label{tab:quantiative_results}
\centering
\resizebox{\textwidth}{!}{
    \small
    \centering
    \setlength{\tabcolsep}{1mm} 
        \begin{threeparttable}
            \begin{tabular}{c c ccc cc cc}
                \toprule
                \textbf{Benchmark}& \textbf{Method} & \multicolumn{3}{c}{\textbf{Object Replacement}} & \multicolumn{2}{c}{\textbf{Object Addition}} & \multicolumn{2}{c}{\textbf{Object Removal}} \\
                \cmidrule(lr){3-5} \cmidrule(lr){6-7} \cmidrule(lr){8-9}
                & & LPIPS$\downarrow$ & CLIP-O$\uparrow$ & CLIP-NO$\uparrow$ &
                LPIPS$\downarrow$ & CLIP-O$\uparrow$ &
                LPIPS$\downarrow$ & CLIP-NO$\uparrow$ \\
                \midrule
                \midrule
    & {P2P}  & \textbf{0.04} & 20.74 & 80.29 & 0.06 & 15.86 & \textbf{0.06} & 79.09 \\
& {DDS}  & 0.06 & 19.47 & 78.1 & \textbf{0.03} & 15.92 & \textbf{0.06} & 76.07 \\
& {Instruct-Pix2Pix}  & 0.14 & 21.89 & 79.24 & 0.1 & 18.68 & 0.13 & 75.89 \\
&{Instruct-Diff}  & 0.14 & 20.21 & 78.2 & 0.1 & 19.08 & 0.1 & 78.6 \\
\textbf{SynO}&{Pix2Pix-Zero}  & 0.12 & 20.17 & 79.4 & 0.27 & 20.9 & 0.19 & 78.92 \\
&{EDICT}  & 0.07 & 19.29 & 79.18 & 0.05 & 16.7 & 0.08 & 77.88 \\
&{MasaCtrl}  & 0.1 & 21.02 & 78.23 & 0.13 & 17.7 & 0.17 & 78.27 \\
&{Plug-and-Play}  & 0.17 & 21.27 & 79.25 & 0.14 & 16.73 & 0.21 & 78.61 \\
&{\textbf{FlexEdit}}  & \textbf{0.04} & \textbf{24.27} & \textbf{81.47} & \textbf{0.03} & \textbf{22.93} & 0.07 & \textbf{80.69} \\

    \midrule

& {P2P}  & 0.19 & 19.42 & 79.26 & 0.19 & 16.24 & 0.25 & 79.84 \\
& {DDS}  & \textbf{0.04} & 20.0 & 79.32 & \textbf{0.03} & 14.88 & \textbf{0.05} & 80.15 \\
& {Instruct-Pix2Pix}  & 0.12 & 21.51 & 79.14 & 0.07 & 18.58 & 0.22 & 79.93 \\
& {Instruct-Diff}  & 0.12 & 21.45 & 79.72 & 0.09 & 18.16 & 0.12 & 80.54 \\
\textbf{PieBenchO}& {Pix2Pix-Zero}  & 0.14 & 20.83 & 80.49 & 0.12 & 18.14 & 0.24 & \textbf{80.98} \\
& {EDICT}  & 0.07 & 19.85 & 80.1 & 0.04 & 15.98 & 0.09 & 80.75 \\
& {MasaCtrl}  & 0.09 & 18.71 & 78.29 & 0.1 & 16.3 & 0.13 & 79.96 \\
& {Plug-and-Play}  & 0.11 & 19.94 & 79.34 & 0.11 & 16.95 & 0.14 & 79.74 \\
& {\textbf{FlexEdit}}  & 0.05 & \textbf{21.72} & \textbf{80.58} & 0.06 & \textbf{19.75} & \textbf{0.05} & 80.89 \\

\midrule

& {P2P}  & 0.23 & 17.27 & 80.83 & 0.27 & 17.84 & 0.27 & \textbf{82.53} \\
 & {DDS}  & \textbf{0.05} & 18.86 & 79.98 & \textbf{0.05} & 18.32 & \textbf{0.06} & 82.21 \\
 &{Instruct-Pix2Pix}  & 0.13 & \textbf{20.9} & 79.13 & 0.13 & 21.06 & 0.12 & 80.83 \\
  &{Instruct-Diff}  & 0.09 & 20.32 & 80.02 & 0.1 & \textbf{21.26} & 0.11 & 81.1 \\
 \textbf{MagicO} & {Pix2Pix-Zero}  & 0.17 & 18.24 & 81.11 & 0.23 & 19.56 & 0.28 & 82.02 \\
& {EDICT}  & 0.07 & 18.38 & 80.51 & 0.08 & 18.78 & 0.09 & 82.33 \\
& {MasaCtrl}  & 0.12 & 15.94 & 78.97 & 0.14 & 17.33 & 0.15 & 81.2 \\
& {Plug-and-Play}  & 0.14 & 17.41 & 79.45 & 0.16 & 17.96 & 0.15 & 82.23 \\
& {\textbf{FlexEdit}}  & 0.07 & 20.58 & \textbf{81.26} & 0.07 & 21.05 & \textbf{0.06} & 82.19 \\

                \bottomrule
            \end{tabular}
        \end{threeparttable}
}
\vspace{-5mm}
\end{table}

\begin{figure}[t]
    \centering
\includegraphics[width=\textwidth]{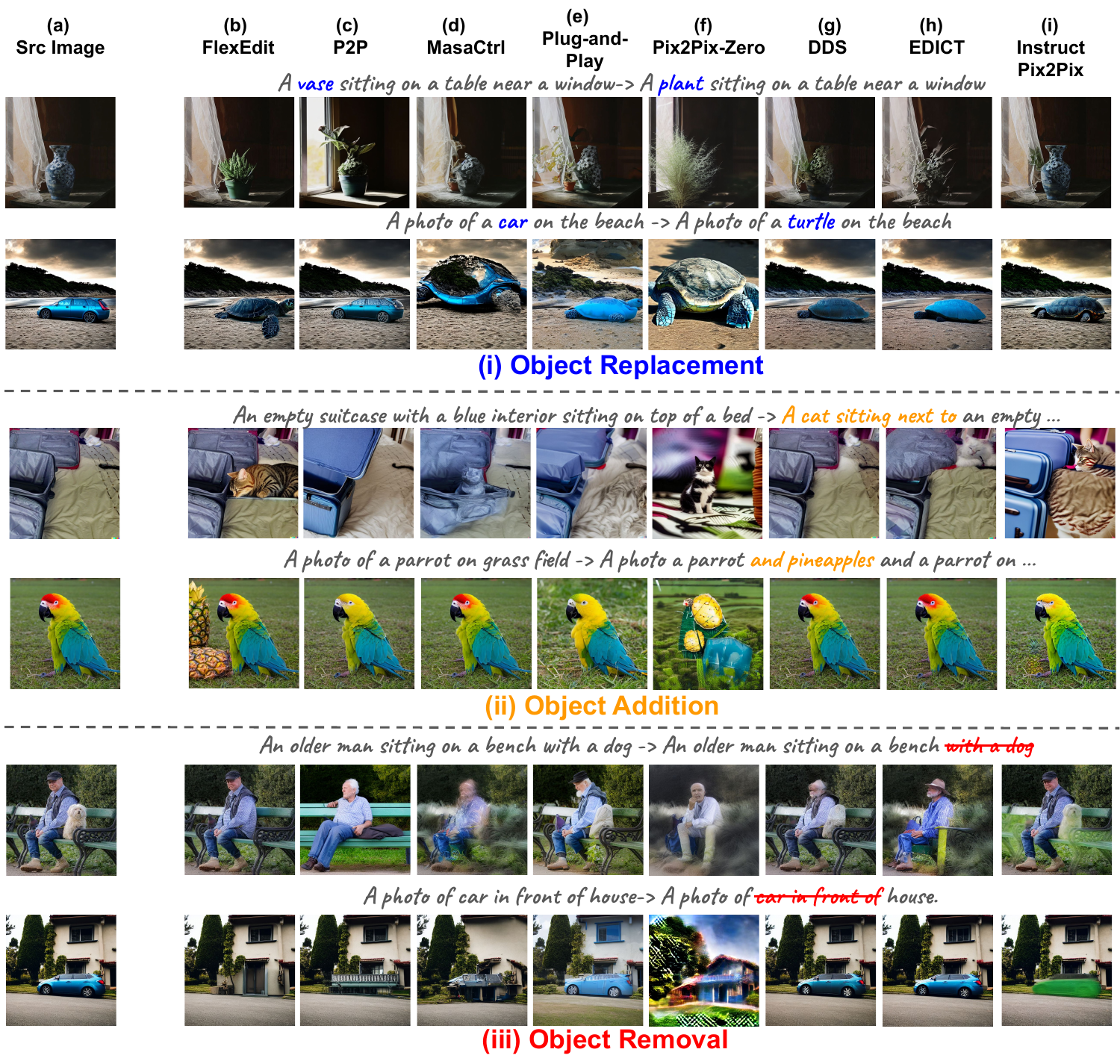}
\vspace{-4mm}
    \caption{ Visualization of comparison results on 3 different editing scenario for different editing methods denoted on top of each column.}
\vspace{-3mm}
\label{fig:all_qualitative_results}
\end{figure}

\begin{figure}[h!]
    \centering
\includegraphics[width=\textwidth]{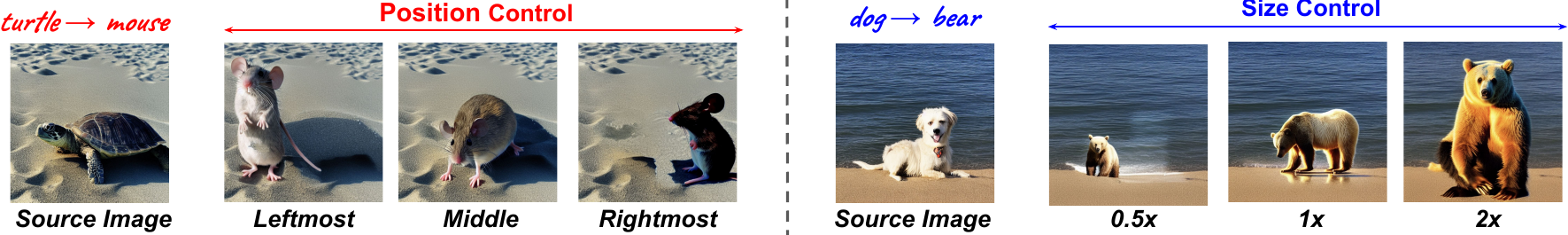}
\vspace{-3mm}
    \caption{Controllable object replacement.}
\vspace{-5mm}
\label{fig:additional_results}
\end{figure}

\minisection{Main Qualitative Results.}
In \cref{fig:all_qualitative_results}, we visualize edited results as the comparison between our FlexEdit and other methods across three editing tasks. For object replacement, our FlexEdit achieves flexible object transformation such as "\textit{car}" to "\textit{turtle}" when others such as DDS, EDICT, or InstructPix2Pix fail to render appropriate texture and shape for "\textit{turtle}". On the other hand, MasaCtrl, Plug-and-Play, or Pix2Pix-Zero could achieve high editing semantics but fail to preserve source background content.
For object addition, except ours, most methods fail to add the new object specified by the text prompt. In the example of "adding pineapples" to the image, P2P and InstructPix2Pix struggle to allocate accurate editing regions, affecting existing objects, i.e., "parrot". Other methods produce results that are mostly identical to the source image, whereas Pix2PixZero destroys the source image's structure.
For object removal, FlexEdit produces satisfactory results in removing the object and inpainting in the missing regions, resulting in natural-looking edited outcomes. Other methods struggle to pinpoint the right object for removal, degrading the edited image's quality.

\minisection{Controllable Editing.} We also demonstrate FlexEdit's capability of controllable editing in \cref{fig:additional_results}. 
FlexEdit achieves reasonable edits by controlling the target object position and size while preserving other source image's details. 

\begin{figure}[t]
    \centering
    \begin{subfigure}[b]{\textwidth}
        \centering
        \includegraphics[width=\textwidth]{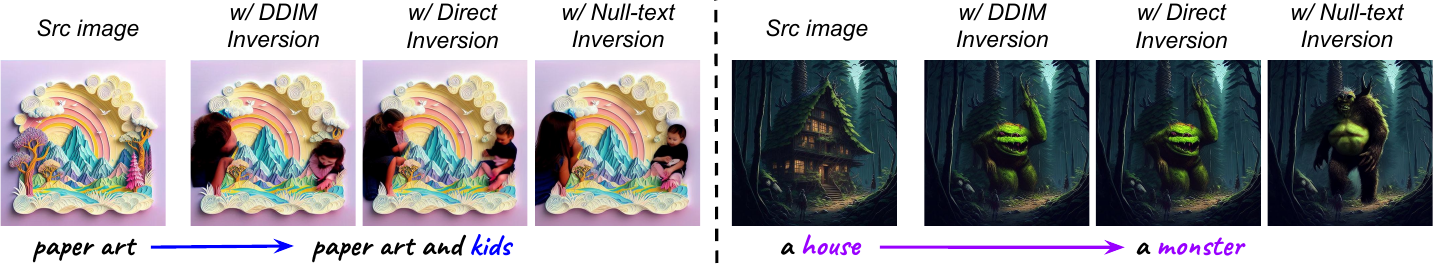}
        \caption{Inversion robustness.} 
        \vspace{1mm}
        \label{fig:inversion_ablate}
    \end{subfigure}
    \begin{subfigure}[b]{\textwidth}
        \centering
        \includegraphics[width=\textwidth]{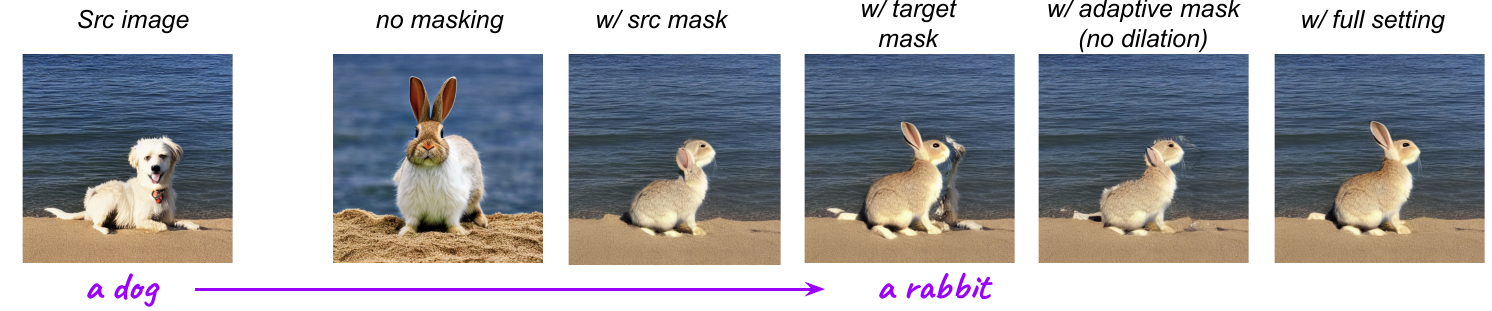}
        \caption{Mask design for latent blending.} 
        \vspace{1mm}
        \label{fig:mask_ablate}
    \end{subfigure}
    \begin{subfigure}[b]{\textwidth}
        \centering
        \includegraphics[width=\textwidth]{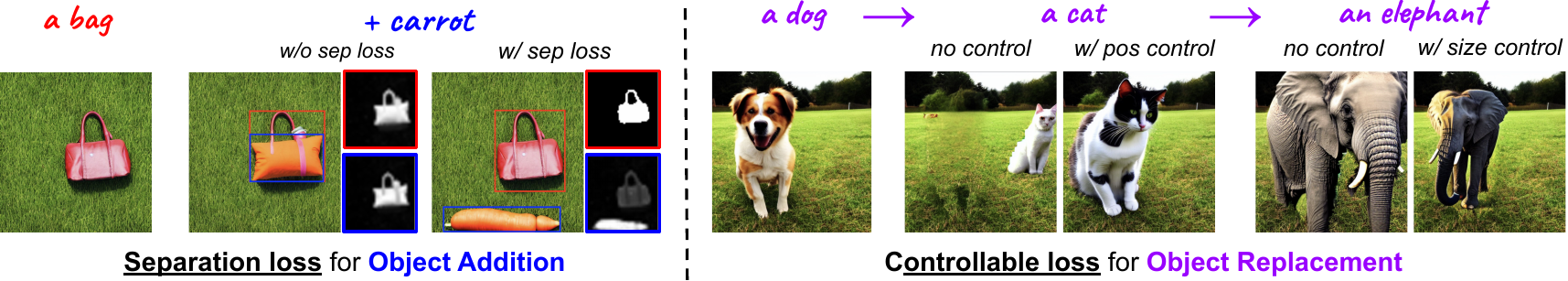}
        \caption{Effect of object constraints.}
\label{fig:optimization_ablate}
    \end{subfigure}
    \vspace{-3mm}
    \caption{Ablation studies. Note that on the left panel of (c), we illustrate the impact of the separation loss by visualizing the editing results along with averaged cross-attention maps for the \textcolor{red}{source} and \textcolor{blue}{added object.}}
    \vspace{-5mm}
    \label{fig:ablations}
\end{figure}

\minisection{Human Preference Study.} In addition, we conducted a human preference study for edited images generated by our FlexEdit compared to other editing methods. We choose DDS\cite{hertz2023delta} and InstructPix2Pix\cite{brooks2023instructpix2pix} as compared methods due to their competitive trade-off scores (\cref{fig:scatter_plot}). The study comprises 60 questions, with 20 questions for each editing task. We ask users to select the best result from three edited images generated from our FlexEdit and two others based on criteria discussed in \cref{sec:evaluation_metrics}. Overall, results from 250 participants show that FlexEdit achieves dominant preference statistics. In object addition, removal, and replacement scenarios, our edited results are preferred by approximately $88.55\%, 96.53\%$, and $88.91\%$ of users for editing semantics and by $68.12\%, 83.95\%$, and $76.72\%$ of users for background preservation.




\subsection{Ablation Studies}
\minisection{Robustness to Inversion Method.}
For diffusion-based real image editing, an inversion method is required to convert the source image into noisy latents. Our framework is not confined to any specific inversion method. As shown in \cref{fig:inversion_ablate}, when combining with any inversion method, 
FlexEdit yields editing results aligned with the specified object in the text prompt. 

    \minisection{Latent Blending with Adaptive Mask.} 
    In \cref{fig:mask_ablate}, we compare editing results using different mask settings in the latent blending mechanism. Without a mask, the output loses content from the source image. Using only the source mask leads to incomplete object generation, while using only the target mask leaves artifacts. Combining both masks resolves these issues but may create visible seams. Our adaptive mask with dilation yields the best results.

\minisection{Loss Constraints.}
We show the effect of our proposed losses in \cref{fig:optimization_ablate}. On the left pane, we show that without separation loss, added objects often blend with existing ones due to unresolved designated regions. Our separation loss helps to separate objects clearly, resulting in more coherent object addition editing. On the right panel, we show that uncontrolled editing leads to randomness in the target object's position and size, and our constraints can mitigate that issue by enforcing the desired properties on the target object.
\section{Conclusions}
We introduce FlexEdit, a novel editing framework that is built upon Stable Diffusion model to enable flexible and controllable object-centric image editing. Our approach employs optimization with object constraints and a latent blending mechanism using an adaptive mask to manipulate latents during denoising. FlexEdit ensures editing semantics while preserving source image fidelity, extensively verified through qualitative and quantitative results across various scenarios and benchmarks. We also introduce a curated evaluation test suite to leverage the object-centric image editing tasks. 

\minisection{Limitations and future work}. While our method achieves a reasonable trade-off between editing semantics and background preservation, it may fail due to incorrect object masks generated by the DINO-SAM model or from attention maps. In addition, the editing process relying on multiple steps of the diffusion model is quite time-consuming. Exploring one-step diffusion model for editing is an interesting direction for our future work. 

\clearpage  

%
%
\bibliographystyle{splncs04}
\bibliography{main}

\begin{thebibliography}{10}
\providecommand{\url}[1]{\texttt{#1}}
\providecommand{\urlprefix}{URL }
\providecommand{\doi}[1]{https://doi.org/#1}

\bibitem{agarwal2023astar}
Agarwal, A., Karanam, S., Joseph, K., Saxena, A., Goswami, K., Srinivasan, B.V.: A-star: Test-time attention segregation and retention for text-to-image synthesis. In: Proceedings of the IEEE/CVF International Conference on Computer Vision. pp. 2283--2293 (2023)

\bibitem{bartal2022text2live}
Bar-Tal, O., Ofri-Amar, D., Fridman, R., Kasten, Y., Dekel, T.: Text2live: Text-driven layered image and video editing. In: European Conference on Computer Vision. pp. 707--723. Springer (2022)

\bibitem{brooks2023instructpix2pix}
Brooks, T., Holynski, A., Efros, A.A.: Instructpix2pix: Learning to follow image editing instructions. In: Proceedings of the IEEE/CVF Conference on Computer Vision and Pattern Recognition. pp. 18392--18402 (2023)

\bibitem{cao2023masactrl}
Cao, M., Wang, X., Qi, Z., Shan, Y., Qie, X., Zheng, Y.: Masactrl: Tuning-free mutual self-attention control for consistent image synthesis and editing. In: {IEEE/CVF} International Conference on Computer Vision, {ICCV} 2023, Paris, France, October 1-6, 2023. pp. 22503--22513. IEEE (2023). \doi{10.1109/ICCV51070.2023.02062}, \url{https://doi.org/10.1109/ICCV51070.2023.02062}

\bibitem{chefer2023attendandexcite}
Chefer, H., Alaluf, Y., Vinker, Y., Wolf, L., Cohen-Or, D.: Attend-and-excite: Attention-based semantic guidance for text-to-image diffusion models. ACM Transactions on Graphics  (2023). \doi{10.1145/3592116}

\bibitem{couairon2022diffedit}
Couairon, G., Verbeek, J., Schwenk, H., Cord, M.: Diffedit: Diffusion-based semantic image editing with mask guidance. In: The Eleventh International Conference on Learning Representations, {ICLR} 2023, Kigali, Rwanda, May 1-5, 2023. OpenReview.net (2023), \url{https://openreview.net/forum?id=3lge0p5o-M-}

\bibitem{epstein2023diffusion}
Epstein, D., Jabri, A., Poole, B., Efros, A.A., Holynski, A.: Diffusion self-guidance for controllable image generation. In: Oh, A., Naumann, T., Globerson, A., Saenko, K., Hardt, M., Levine, S. (eds.) Advances in Neural Information Processing Systems 36: Annual Conference on Neural Information Processing Systems 2023, NeurIPS 2023, New Orleans, LA, USA, December 10 - 16, 2023 (2023), \url{http://papers.nips.cc/paper\_files/paper/2023/hash/3469b211b829b39d2b0cfd3b880a869c-Abstract-Conference.html}

\bibitem{geng2023instructdiffusion}
Geng, Z., Yang, B., Hang, T., Li, C., Gu, S., Zhang, T., Bao, J., Zhang, Z., Li, H., Hu, H., et~al.: Instructdiffusion: A generalist modeling interface for vision tasks. In: Proceedings of the IEEE/CVF Conference on Computer Vision and Pattern Recognition. pp. 12709--12720 (2024)

\bibitem{hertz2023delta}
Hertz, A., Aberman, K., Cohen-Or, D.: Delta denoising score. In: Proceedings of the IEEE/CVF International Conference on Computer Vision. pp. 2328--2337 (2023)

\bibitem{hertz2022prompttoprompt}
Hertz, A., Mokady, R., Tenenbaum, J., Aberman, K., Pritch, Y., Cohen-Or, D.: Prompt-to-prompt image editing with cross attention control. International Conference on Learning Representations  (2022). \doi{10.48550/arXiv.2208.01626}

\bibitem{DBLP:journals/corr/abs-2009-00726}
Hu, X., Zhang, Z., Jiang, Z., Chaudhuri, S., Yang, Z., Nevatia, R.: Span: Spatial pyramid attention network for image manipulation localization. In: Computer Vision--ECCV 2020: 16th European Conference, Glasgow, UK, August 23--28, 2020, Proceedings, Part XXI 16. pp. 312--328. Springer (2020)

\bibitem{ju2023direct}
Ju, X., Zeng, A., Bian, Y., Liu, S., Xu, Q.: Pnp inversion: Boosting diffusion-based editing with 3 lines of code. In: The Twelfth International Conference on Learning Representations (2024)

\bibitem{kawar2023imagic}
Kawar, B., Zada, S., Lang, O., Tov, O., Chang, H., Dekel, T., Mosseri, I., Irani, M.: Imagic: Text-based real image editing with diffusion models. In: Proceedings of the IEEE/CVF Conference on Computer Vision and Pattern Recognition. pp. 6007--6017 (2023)

\bibitem{li2023divide}
Li, Y., Keuper, M., Zhang, D., Khoreva, A.: Divide \& bind your attention for improved generative semantic nursing. In: 34th British Machine Vision Conference 2023, {BMVC} 2023 (2023)

\bibitem{meng2022sdedit}
Meng, C., He, Y., Song, Y., Song, J., Wu, J., Zhu, J.Y., Ermon, S.: Sdedit: Guided image synthesis and editing with stochastic differential equations. International Conference on Learning Representations  (2021)

\bibitem{mirzaei2023watch}
Mirzaei, A., Aumentado-Armstrong, T., Brubaker, M.A., Kelly, J., Levinshtein, A., Derpanis, K.G., Gilitschenski, I.: Watch your steps: Local image and scene editing by text instructions. In: ECCV (2024)

\bibitem{mokady2022nulltext}
Mokady, R., Hertz, A., Aberman, K., Pritch, Y., Cohen{-}Or, D.: Null-text inversion for editing real images using guided diffusion models. In: {IEEE/CVF} Conference on Computer Vision and Pattern Recognition, {CVPR} 2023, Vancouver, BC, Canada, June 17-24, 2023. pp. 6038--6047. IEEE (2023). \doi{10.1109/CVPR52729.2023.00585}, \url{https://doi.org/10.1109/CVPR52729.2023.00585}

\bibitem{nguyen2023dataset}
Nguyen, Q.H., Vu, T.T., Tran, A.T., Nguyen, K.: Dataset diffusion: Diffusion-based synthetic data generation for pixel-level semantic segmentation. In: Thirty-seventh Conference on Neural Information Processing Systems (2023), \url{https://openreview.net/forum?id=StD4J5ZlI5}

\bibitem{parmar2023zeroshot}
Parmar, G., Kumar~Singh, K., Zhang, R., Li, Y., Lu, J., Zhu, J.Y.: Zero-shot image-to-image translation. In: ACM SIGGRAPH 2023 Conference Proceedings. pp. 1--11 (2023)

\bibitem{radford2021learning}
Radford, A., Kim, J.W., Hallacy, C., Ramesh, A., Goh, G., Agarwal, S., Sastry, G., Askell, A., Mishkin, P., Clark, J., et~al.: Learning transferable visual models from natural language supervision. In: International conference on machine learning. pp. 8748--8763. PMLR (2021)

\bibitem{ramesh2022hierarchical}
Ramesh, A., Dhariwal, P., Nichol, A., Chu, C., Chen, M.: Hierarchical text-conditional image generation with clip latents. arXiv preprint arXiv:2204.06125  \textbf{1}(2), ~3 (2022)

\bibitem{ren2024grounded}
Ren, T., Liu, S., Zeng, A., Lin, J., Li, K., Cao, H., Chen, J., Huang, X., Chen, Y., Yan, F., Zeng, Z., Zhang, H., Li, F., Yang, J., Li, H., Jiang, Q., Zhang, L.: Grounded sam: Assembling open-world models for diverse visual tasks. ArXiv  \textbf{abs/2401.14159} (2024), \url{https://api.semanticscholar.org/CorpusID:267212047}

\bibitem{rombach2022highresolution}
Rombach, R., Blattmann, A., Lorenz, D., Esser, P., Ommer, B.: High-resolution image synthesis with latent diffusion models. In: Proceedings of the IEEE/CVF conference on computer vision and pattern recognition. pp. 10684--10695 (2022)

\bibitem{saharia2022photorealistic}
Saharia, C., Chan, W., Saxena, S., Li, L., Whang, J., Denton, E.L., Ghasemipour, S.K.S., Lopes, R.G., Ayan, B.K., Salimans, T., Ho, J., Fleet, D.J., Norouzi, M.: Photorealistic text-to-image diffusion models with deep language understanding. In: Koyejo, S., Mohamed, S., Agarwal, A., Belgrave, D., Cho, K., Oh, A. (eds.) Advances in Neural Information Processing Systems 35: Annual Conference on Neural Information Processing Systems 2022, NeurIPS 2022, New Orleans, LA, USA, November 28 - December 9, 2022 (2022), \url{http://papers.nips.cc/paper\_files/paper/2022/hash/ec795aeadae0b7d230fa35cbaf04c041-Abstract-Conference.html}

\bibitem{song2022denoising}
Song, J., Meng, C., Ermon, S.: Denoising diffusion implicit models. International Conference on Learning Representations  (2020)

\bibitem{tumanyan2022plugandplay}
Tumanyan, N., Geyer, M., Bagon, S., Dekel, T.: Plug-and-play diffusion features for text-driven image-to-image translation. In: Proceedings of the IEEE/CVF Conference on Computer Vision and Pattern Recognition. pp. 1921--1930 (2023)

\bibitem{tunstall2023zephyr}
Tunstall, L., Beeching, E., Lambert, N., Rajani, N., Rasul, K., Belkada, Y., Huang, S., von Werra, L., Fourrier, C., Habib, N., Sarrazin, N., Sanseviero, O., Rush, A.M., Wolf, T.: Zephyr: Direct distillation of lm alignment. arXiv preprint arXiv: 2310.16944  (2023)

\bibitem{valevski2023unitune}
Valevski, D., Kalman, M., Molad, E., Segalis, E., Matias, Y., Leviathan, Y.: Unitune: Text-driven image editing by fine tuning a diffusion model on a single image. {ACM} Trans. Graph.  \textbf{42}(4),  128:1--128:10 (2023). \doi{10.1145/3592451}, \url{https://doi.org/10.1145/3592451}

\bibitem{wallace2022edict}
Wallace, B., Gokul, A., Naik, N.: Edict: Exact diffusion inversion via coupled transformations. In: Proceedings of the IEEE/CVF Conference on Computer Vision and Pattern Recognition. pp. 22532--22541 (2023)

\bibitem{wang2022objectformer}
Wang, J., Wu, Z., Chen, J., Han, X., Shrivastava, A., Lim, S.N., Jiang, Y.G.: Objectformer for image manipulation detection and localization. In: Proceedings of the IEEE/CVF Conference on Computer Vision and Pattern Recognition. pp. 2364--2373 (2022)

\bibitem{8953774}
Wu, Y., AbdAlmageed, W., Natarajan, P.: Mantra-net: Manipulation tracing network for detection and localization of image forgeries with anomalous features. In: 2019 IEEE/CVF Conference on Computer Vision and Pattern Recognition (CVPR). pp. 9535--9544 (2019). \doi{10.1109/CVPR.2019.00977}

\bibitem{xie2023boxdiff}
Xie, J., Li, Y., Huang, Y., Liu, H., Zhang, W., Zheng, Y., Shou, M.Z.: Boxdiff: Text-to-image synthesis with training-free box-constrained diffusion. In: Proceedings of the IEEE/CVF International Conference on Computer Vision. pp. 7452--7461 (2023)

\bibitem{zhang2023magicbrush}
Zhang, K., Mo, L., Chen, W., Sun, H., Su, Y.: Magicbrush: A manually annotated dataset for instruction-guided image editing. Neural Information Processing Systems  (2023). \doi{10.48550/arXiv.2306.10012}

\bibitem{zhang2022sine}
Zhang, Z., Han, L., Ghosh, A., Metaxas, D.N., Ren, J.: Sine: Single image editing with text-to-image diffusion models. In: Proceedings of the IEEE/CVF Conference on Computer Vision and Pattern Recognition. pp. 6027--6037 (2023)

\bibitem{zhou2018learning}
Zhou, P., Han, X., Morariu, V.I., Davis, L.S.: Learning rich features for image manipulation detection. In: Proceedings of the IEEE conference on computer vision and pattern recognition. pp. 1053--1061 (2018)

\end{thebibliography}

\clearpage
\appendix

\noindent\textbf{\Large{Appendix}}
\vspace{0.5cm}

In this supplementary material, we first demonstrate how we construct object-centric editing benchmarks (MagicO, PiebenchO, and SynO) in \cref{sec:benchmark_construction}. We then show in detail how we implement the baselines used for comparison in \cref{sec:implemention_details}. We also provide additional quantitative and qualitative results of FlexEdit compared to other editing approaches in \cref{sec:additional_results}. In \cref{sec:scocial_impact}, we discuss the societal impact of FlexEdit. Finally, we demonstrate the potential application of FlexEdit for various editing scenarios in \cref{sec:editing_applications}.
\section{Benchmark Construction}
\label{sec:benchmark_construction}

\subsection{Synthesized Image Editing}
For synthesized image editing, we design editing prompts and instructions to capture object-centric editing scenarios, including object replacement, addition, and removal. Each editing sample consists of a source prompt, a target prompt, the original synthesized image generated by the source prompt, and an editing instruction (for comparison with instruction-based image editing baselines). We utilize Stable Diffusion v1.4 as the generative model to synthesize the source image using the source prompt. For both editing prompt and instruction, we develop several templates covering all three editing tasks: object replacement, addition, and removal, as follows:
\begin{lstlisting}[style=myverbatim]
# Template of generated prompts and instructions for SynO.
## Object Replacement 
Source Prompt: "A photo of <A_OBJECT> <C_BACKGROUND>."
Target Prompt: "A photo of <B_OBJECT> <C_BACKGROUND>."
Instruction: "Turn <A_OBJECT> into <B_OBJECT>."

## Object Addition 
Source Prompt: "A photo of <A_OBJECT> <C_BACKGROUND>."
Target Prompt: "A photo of <A_OBJECT> and <B_OBJECT> <C_BACKGROUND>."
Instruction: "Add <B_OBJECT> next to <A_OBJECT>"

## Object Removal 
Source Prompt: "A photo of <A_OBJECT> <C_BACKGROUND>."
Target Prompt: "A photo of <C_BACKGROUND>."
Instruction: "Remove <A_OBJECT>."
\end{lstlisting}

We designate \lstinline[style=myverbatim]{<A_OBJECT>} as the source object, \lstinline[style=myverbatim]{<B_OBJECT>} as another object used for editing operation, and \lstinline[style=myverbatim]{<C_BACKGROUND>} as the selected background corresponding to the object. To complete prompts and instructions using the above template, we design several object groups along with corresponding backgrounds to fill in the tokens \lstinline[style=myverbatim]{<A_OBJECT>}, \lstinline[style=myverbatim]{<B_OBJECT>}, and \lstinline[style=myverbatim]{<C_BACKGROUND>}. Each group of objects includes a list of objects that are semantically similar to each other, along with a list of background contexts that capture object in real-world scenarios. The list of groups is constructed as follows:
\begin{lstlisting}[style=myverbatim]
{
    "group1": {
        "name": "animal",
        "list_objects": ["parrot", "monkey", "bird", "turtle", "cat", "dog", "elephant", "giraffe", "lion", "horse", "bear"],
        "background": ["on beach", "on grass field", "in the forest", "on street"]
    },
    "group2": {
        "name": "transportation",
        "list_objects": ["car", "bicycle", "boat"],
        "background": ["on street", "on beach", "in front of house"]
    },
    "group3": {
        "name": "fruit",
        "list_objects": ["apple", "banana", "orange", "avocado", "pineapple", "pear"],
        "background": ["on table", "hanging on tree", "on grass field"]
    },
    "group4":{
        "name": "furniture",
        "list_objects": ["chair", "table", "sofa"],
        "background": ["in living room", "in the kitchen"]
    },
    "group5":{
        "name": "musical instruments",
        "list_objects": ["guitar", "piano", "violin", "drums"],
        "background": ["on grass field", "on table"]
    },
    "group6":{
        "name": "household appliances",
        "list_objects": ["refrigerator", "microwave", "toaster"],
        "background": ["in living room", "in the kitchen"]
    }
}
\end{lstlisting}

\subsection{Real-Image Editing Benchmark}

To construct MagicO and PieBenchO, we utilize the Magic Brush\cite{zhang2023magicbrush} and Piebench\cite{ju2023direct} test suites to curate editing samples relevant to object-centric editing problems. Specifically, we provide an input prompt wrapped with an editing instruction to the language model Zephyr \cite{tunstall2023zephyr} to determine whether a corresponding editing sample relates to object replacement, addition, or removal. The prompt is constructed for each editing scenario and includes a few input-output pairs as examples to help the language model better capture the context of the prompt. The prompts used to construct both MagicO and PiebenchO for individual object editing scenarios, such as object replacement, addition, and removal, are shown in \cref{tab:prompt_real_edit}. We then curate images marked as relevant editing samples returned by Zephyr based on the provided annotation JSON file from both MagicBrush and PieBench test suites to construct MagicO and PiebenchO.  
\section{Implementation Details}
\label{sec:implemention_details}

For all editing approaches built upon Stable Diffusion\cite{rombach2022highresolution} (SD), we employ the same Stable Diffusion v1.4 with similar model's default hyperparemeters: $T=50$ for number of diffusion steps, and $w=7.5$ for guidance scale. When performing synthesis image editing, we keep track of the intermediate latents during the denoising process to generate synthesized images. These intermediate latents are then utilized similarly for all editing approaches which rely on an inversion process. 

For quantitative and qualitative comparisons shown in the main paper, we use the official implementation of P2P \cite{hertz2022prompttoprompt}, MasaCtrl \cite{cao2023masactrl}, Plug-and-Play \cite{tumanyan2022plugandplay}, Pix2Pix-Zero \cite{parmar2023zeroshot}, EDICT\cite{wallace2022edict}, DDS\cite{hertz2023delta}, InstructPix2Pix\cite{brooks2023instructpix2pix}, and InstructDiffusion\cite{geng2023instructdiffusion}. For inversion techniques, we also use the official implementation of Direct-Inversion \cite{ju2023direct} and Null-text Inversion\cite{mokady2022nulltext} with their default hyper-parameters. All the experiments are conducted on a single NVIDIA V100 GPU.

\section{Additional Results}
\label{sec:additional_results}
\subsection{Quantiative Results}
\subsubsection{Other Inversion Methods.}
In addition to the comparison results presented in the main paper, we also provide supplementary comparison results for editing methods that rely on an inversion process. Specifically, we apply other inversion methods such as Direct Inversion \cite{ju2023direct} on editing methods such as MasaCtrl\cite{cao2023masactrl}, Plug-and-Play\cite{tumanyan2022plugandplay}, and Pix2Pix-Zero \cite{parmar2023zeroshot} and Null-text Inversion\cite{mokady2022nulltext} on P2P\cite{hertz2022prompttoprompt}. Our FlexEdit is not limited to any particular inversion method and can be combined with both Direct Inversion and Null-text Inversion. In \cref{tab:direct_inversion_compare} and \cref{tab:null_text_inversion_compare}, we show the comparison results on different benchmarks and editing tasks using different inversion methods.

The results demonstrate that our FlexEdit consistently outperforms most of the editing approaches using both inversion methods: Direct Inversion and Null-text Inversion, in terms of both background preservation and editing semantics criteria across various benchmarks and editing tasks. We also achieve a reliable trade-off, while other approaches can only maintain good results in one aspect.

\begin{table}[t]
\caption{
        Quantitative comparison of FlexEdit with other editing methods with \textbf{Direct Inversion} across two benchmarks and three editing tasks.
    }
\vspace{-3mm}
\label{tab:direct_inversion_compare}
\centering
\resizebox{\textwidth}{!}{
    \small
    \centering
    \setlength{\tabcolsep}{1mm} 
        \begin{threeparttable}
            \begin{tabular}{c c ccc cc cc}
                \toprule
                \textbf{Benchmark}& \textbf{Method} & \multicolumn{3}{c}{\textbf{Object Replacement}} & \multicolumn{2}{c}{\textbf{Object Addition}} & \multicolumn{2}{c}{\textbf{Object Removal}} \\
                \cmidrule(lr){3-5} \cmidrule(lr){6-7} \cmidrule(lr){8-9}
                & & LPIPS$\downarrow$ & CLIP-O$\uparrow$ & CLIP-NO$\uparrow$ &
                LPIPS$\downarrow$ & CLIP-O$\uparrow$ &
                LPIPS$\downarrow$ & CLIP-NO$\uparrow$ \\
                \midrule

& {P2P}  & 0.09 & 17.80 & 79.86 & 0.17 & 19.16 & 0.12 & 81.97 \\
& {MasaCtrl}  & 0.09 & 16.63 & 78.68 & 0.11 & 17.67 & 0.12 & 81.13 \\
\textbf{MagicO} & {Plug-and-Play}  & 0.11 & 17.53 & 79.5 & 0.13 & 17.82 & 0.14 & \textbf{82.72} \\
& {Pix2Pix-Zero}  & 0.13 & 18.78 & 80.99 & 0.18 & 19.44 & 0.24 & 82.18 \\
& {\textbf{FlexEdit}}  & \textbf{0.07} & \textbf{20.35} & \textbf{81.62} & \textbf{0.07} & \textbf{21.11} & \textbf{0.07} & 82.35 \\

\midrule

& {P2P}  & 0.07 & 20.34 & 79.49 & 0.13 & 17.45 & 0.10 & 79.89 \\
& {Pix2Pix-Zero}  & 0.11 & 20.69 & \textbf{80.49} & 0.09 & 17.78 & 0.20 & \textbf{80.97} \\
\textbf{PieBenchO} & {MasaCtrl}  & 0.08 & 19.05 & 78.08 & 0.08 & 16.25 & 0.10 & 79.75 \\
& {Plug-and-Play}  & 0.10 & 19.88 & 79.33 & 0.10 & 16.64 & 0.13 & 79.58 \\
& {\textbf{FlexEdit}}  & \textbf{0.05} & \textbf{21.78} & 80.43 & \textbf{0.06} & \textbf{19.92} & \textbf{0.05} & 80.80 \\

        \bottomrule
    \end{tabular}
\end{threeparttable}
}
\end{table}

\begin{table}[t]
\caption{
        Quantitative comparison of FlexEdit with other editing methods with \textbf{Null-text Inversion} across two benchmarks and three editing tasks.
    }
\vspace{-3mm}
\label{tab:null_text_inversion_compare}
\centering
\resizebox{\textwidth}{!}{
    \small
    \centering
    \setlength{\tabcolsep}{1mm} 
        \begin{threeparttable}
            \begin{tabular}{c c ccc cc cc}
                \toprule
                \textbf{Benchmark}& \textbf{Method} & \multicolumn{3}{c}{\textbf{Object Replacement}} & \multicolumn{2}{c}{\textbf{Object Addition}} & \multicolumn{2}{c}{\textbf{Object Removal}} \\
                \cmidrule(lr){3-5} \cmidrule(lr){6-7} \cmidrule(lr){8-9}
                & & LPIPS$\downarrow$ & CLIP-O$\uparrow$ & CLIP-NO$\uparrow$ &
                LPIPS$\downarrow$ & CLIP-O$\uparrow$ &
                LPIPS$\downarrow$ & CLIP-NO$\uparrow$ \\
                \midrule

\textbf{MagicO} & {P2P} & \textbf{0.05} & 18.21 & 80.51 & 0.14 & 20.03 & 0.10 & 82.31 \\
 & \textbf{FlexEdit} & 0.07 & \textbf{20.36} & \textbf{81.70} & \textbf{0.08} & \textbf{21.12} & \textbf{0.07} & \textbf{83.26} \\
\midrule
\textbf{PiebenchO} & {P2P} & \textbf{0.05} & 20.28 & 80.57 & 0.09 & 18.08 & 0.07 & 80.83 \\
 & \textbf{FlexEdit} & 0.06 & \textbf{21.11} & \textbf{80.83} & \textbf{0.07} & \textbf{20.38} & \textbf{0.05} & \textbf{81.74} \\

                \bottomrule
            \end{tabular}
        \end{threeparttable}
}
\end{table}


\subsubsection{Inference time comparison.}
We also provide estimated inference time of each editing method including our FlexEdit executed on the same machine setting shown in \cref{tab:inference_time}
\begin{table}[h!]
\centering
\caption{Inference time of different editing techniques}
\label{tab:inference_time}
\begin{adjustbox}{width=\textwidth}
\begin{tabular}{lccccccccc}
\toprule
\multicolumn{1}{l}{\textbf{Methods}} & FlexEdit & MasaCtrl & Plug-and-Play & P2P    & Pix2Pix-Zero & DDS     & EDICT   & Instruct-Pix2Pix & Instruct-Diffusion \\ \midrule
\textbf{Inference time (s)}          & 41.481   & 29.051   & 16.565        & 42.202 & 66.297       & 124.048 & 234.970 & 17.847          & 42.310            \\ \bottomrule
\end{tabular}
\end{adjustbox}
\end{table}
\subsection{Qualitative Results}
\begin{figure}
    \centering
    \includegraphics[width=\textwidth]{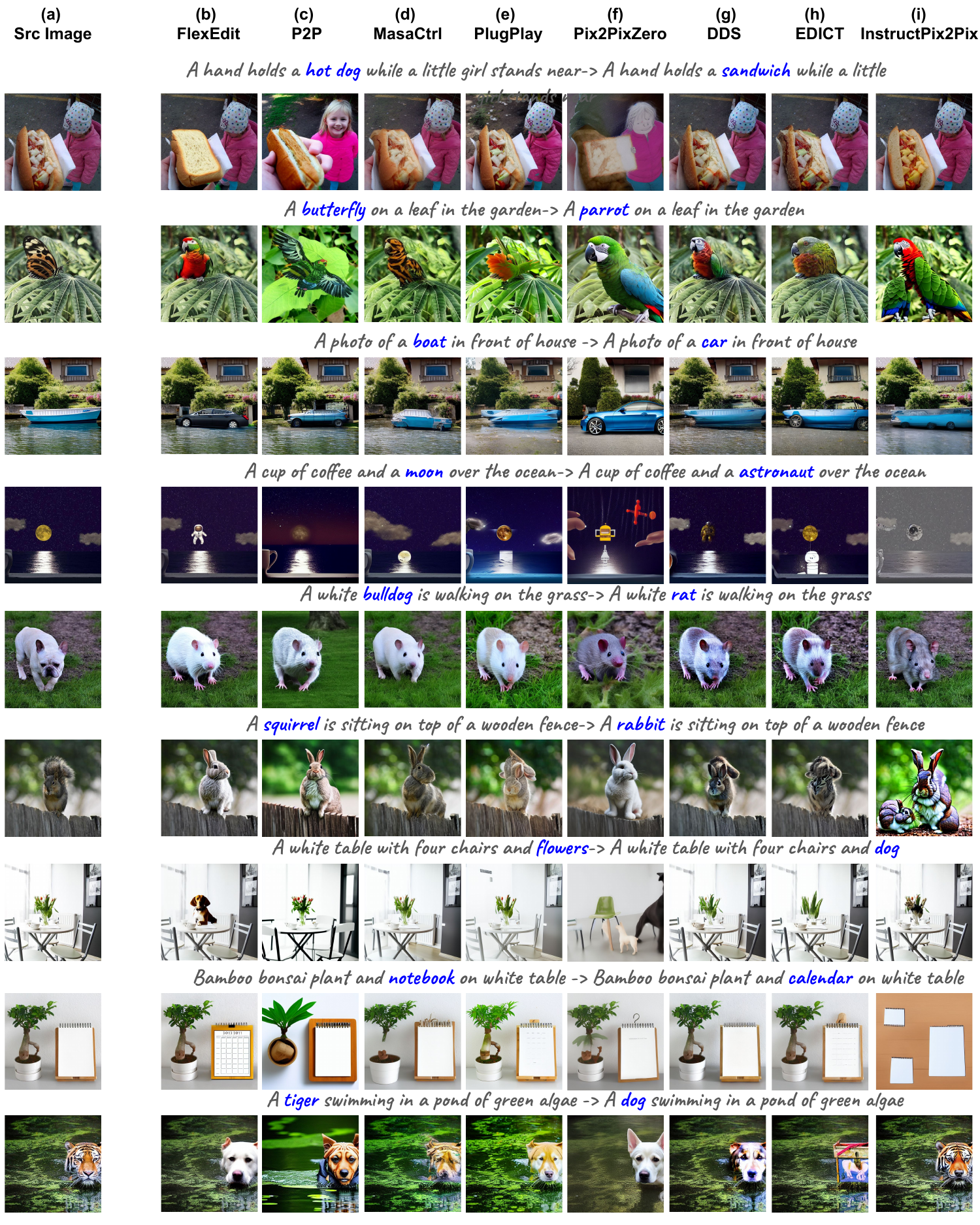}
    \caption{Visualization for comparison of editing results in object replacement task.}
    \label{fig:replace_additional}
\end{figure}

\begin{figure}
    \centering
    \includegraphics[width=\textwidth]{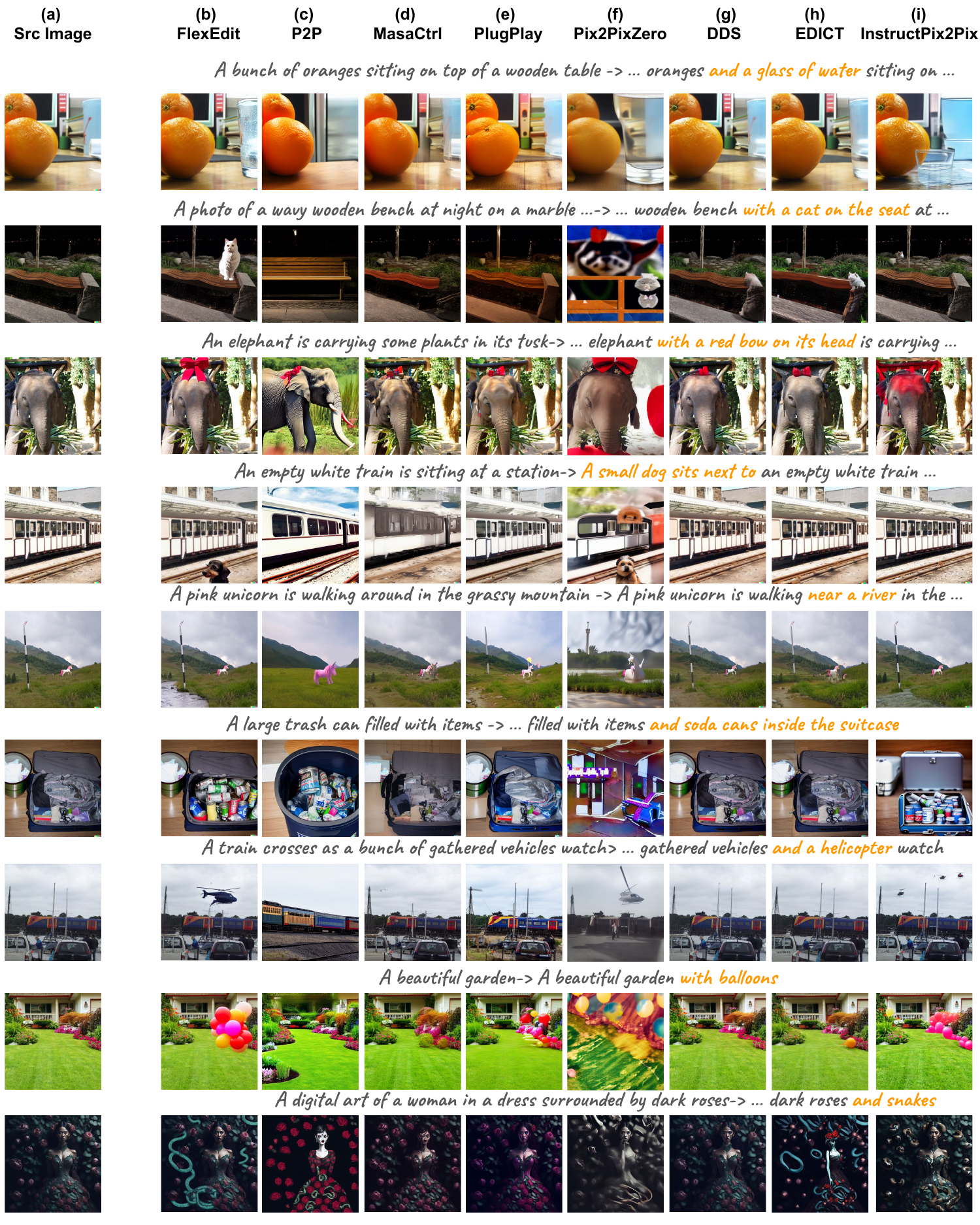}
    \caption{Visualization for comparison of editing results in object addition task.}
    \label{fig:insert_additional}
\end{figure}

\begin{figure}
    \centering
    \includegraphics[width=\textwidth]{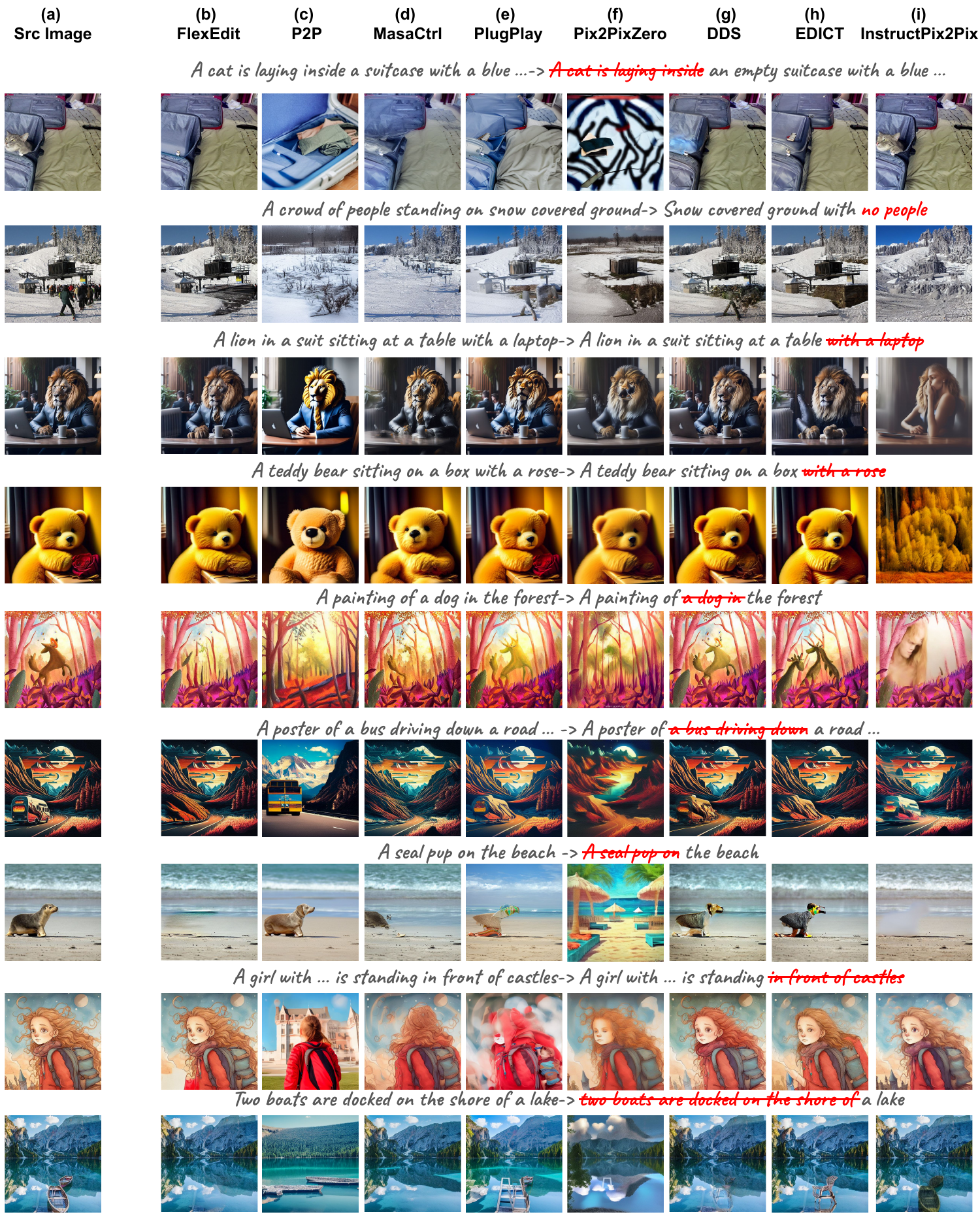}
    \caption{Visualization for comparison of editing results in object removal task.}
    \label{fig:remove_additional}
\end{figure}

Additional editing results of FlexEdit compared with other methods on different benchmarks are provided in \cref{fig:replace_additional}, \cref{fig:insert_additional}, and \cref{fig:remove_additional} for three corresponding tasks: object replacement, object addition, and object removal. 

\section{Societal Impacts of FlexEdit}
\label{sec:scocial_impact}
FlexEdit is built upon state-of-the-art text-to-image diffusion models, which enable image manipulation to achieve desirable edits. As an AI-powered visual generation tool, FlexEdit offers customizable content generation capabilities to the visual art community. By automating image editing tasks, FlexEdit enhances the efficiency of various visual creation endeavors. We also recognize the ethical and societal challenges that come with the widespread adoption of our editing framework. FlexEdit could be exploited by malicious parties to produce sensitive or unrealistic content aiming to spread disinformation. We believe these issues should be addressed, and we can thoroughly engineer its capabilities to fulfill their intended functions while possessing the ability to detect and avoid unintended consequences and behavior in the future. Several ongoing research works have been conducted aiming to mitigate such issues, including detecting and localizing image manipulation\cite{wang2022objectformer, 8953774, zhou2018learning, DBLP:journals/corr/abs-2009-00726}.

\begin{table*}[t]
\setlength\tabcolsep{0pt}
\caption{Our full prompt for prompting Zephyr model to extract relevant editing information for constructing MagicO and PiebenchO.}
\centering
\begin{tabular*}{\linewidth}{@{\extracolsep{\fill}} l }\toprule
\label{tab:prompt_real_edit}
\begin{lstlisting}[style=myverbatim]
# Your Role: You are a friendly chatbot who always responds in the style of programmer

## Scenario 1: Prompt designed to extract relevant editing information given editing instruction for object replacement.

- User Prompt: "Given the Instruction: {instruction} for object replacement in image editing task. Return in the following string format without any further explanation: A-B where A is the source object and B is the target object."


## Scenario 2: Prompt designed to extract relevant editing information given editing instruction for object removal.

- User prompt: "Given the Instruction: `{instruction}' for object removal in image editing task. Return in the following string format without any further explanation: A-B where A is the source object to be removed and B is None."

## Scenario 3: Prompt designed to extract relevant editing information given editing instruction for object addition.

- User prompt: "Given the Instruction: {instruction} for object adding in image editing task. Return in the following string format without any further explanation: A-B where A is the new object being added, and B is the specified position of where to add object, if there is no position being mentioned, B is None."

\end{lstlisting} \\\bottomrule
\end{tabular*}
\end{table*}

\section{Other Editing Applications}
\label{sec:editing_applications}
\subsection{Flexible Object Replacement}
\begin{figure}
    \centering
    \includegraphics[width=\textwidth]{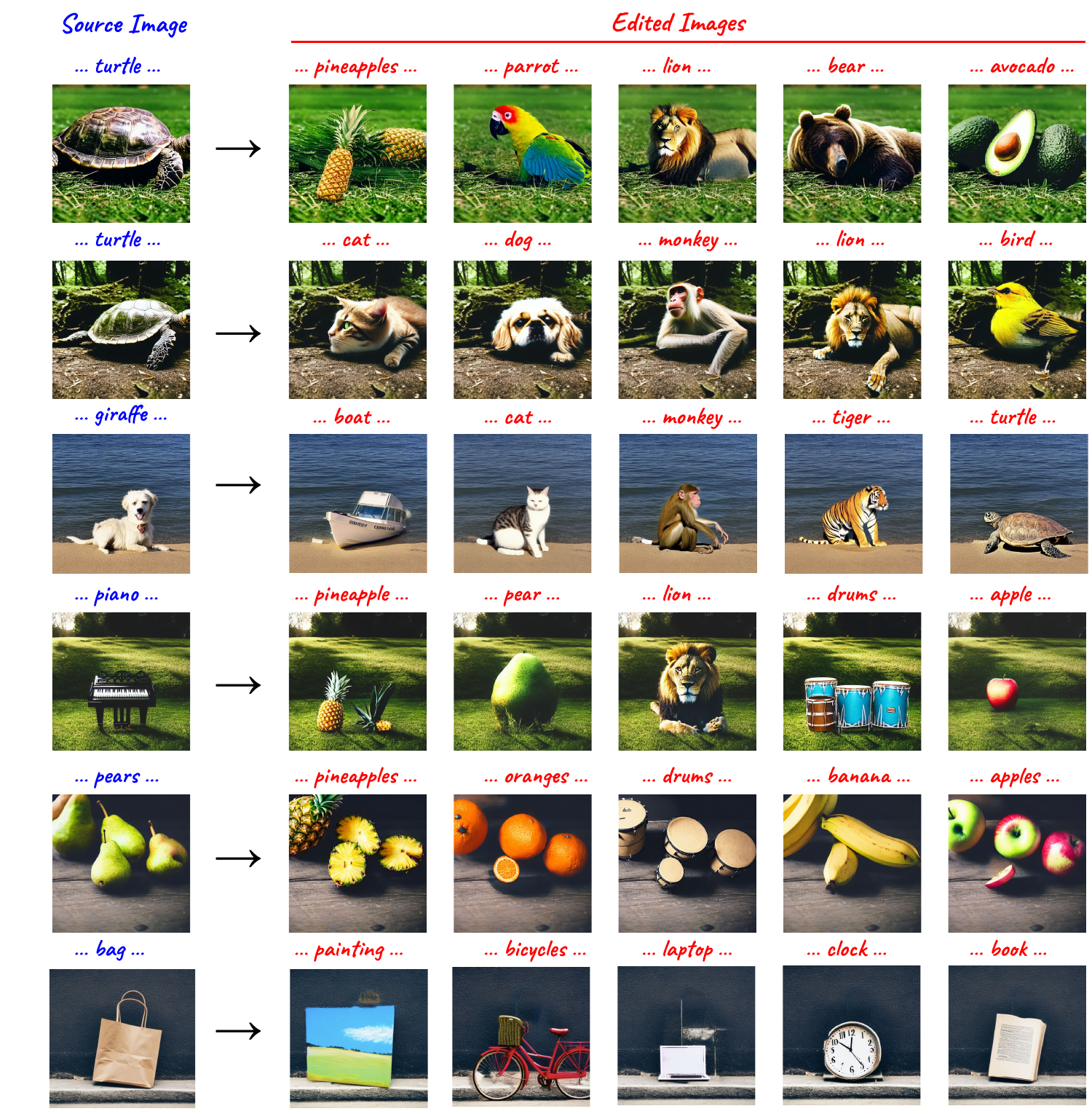}
    \caption{Visualization results of FlexEdit in achieving flexible shape transformation for \textbf{synthesis} object replacement.}
    \label{fig:flex_transformation_syn}
\end{figure}

\begin{figure}
    \centering
    \includegraphics[width=\textwidth]{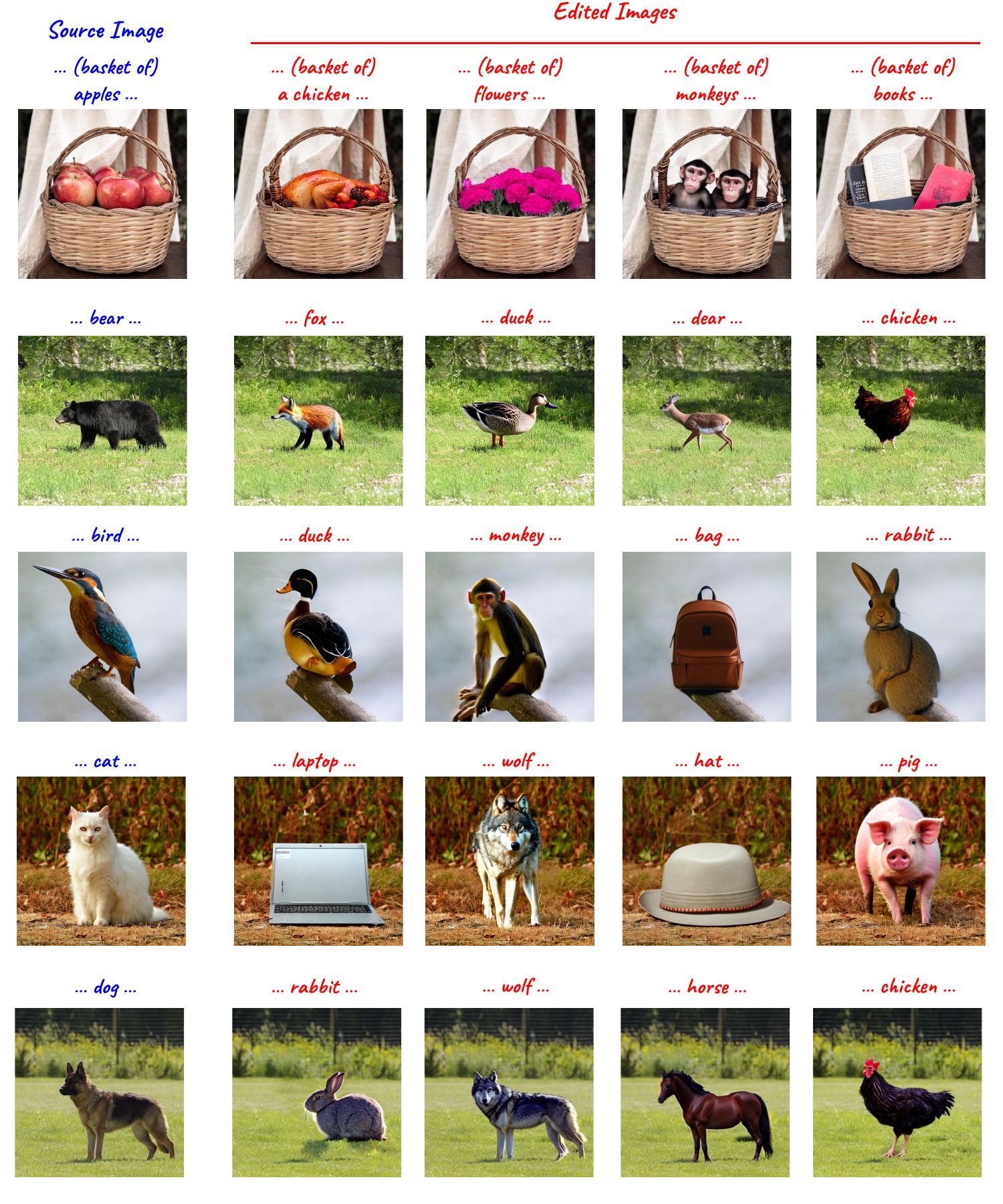}
    \caption{Visualization results of FlexEdit in achieving flexible shape transformation for \textbf{real} object replacement.}
    \label{fig:flex_transformation_real}
\end{figure}

We demonstrate the potential editing application of our FlexEdit framework when achieving flexible shape transformation in object replacement for both synthesized image and real image as shown in \cref{fig:flex_transformation_syn} and \cref{fig:flex_transformation_real}, respectively. As shown in the visualization results, source objects could be edited to transform into several object types with a large variation in terms of shape using our FlexEdit framework.

\subsection{Controllable Object Replacement}
\begin{figure}
    \centering
    \includegraphics[width=\textwidth]{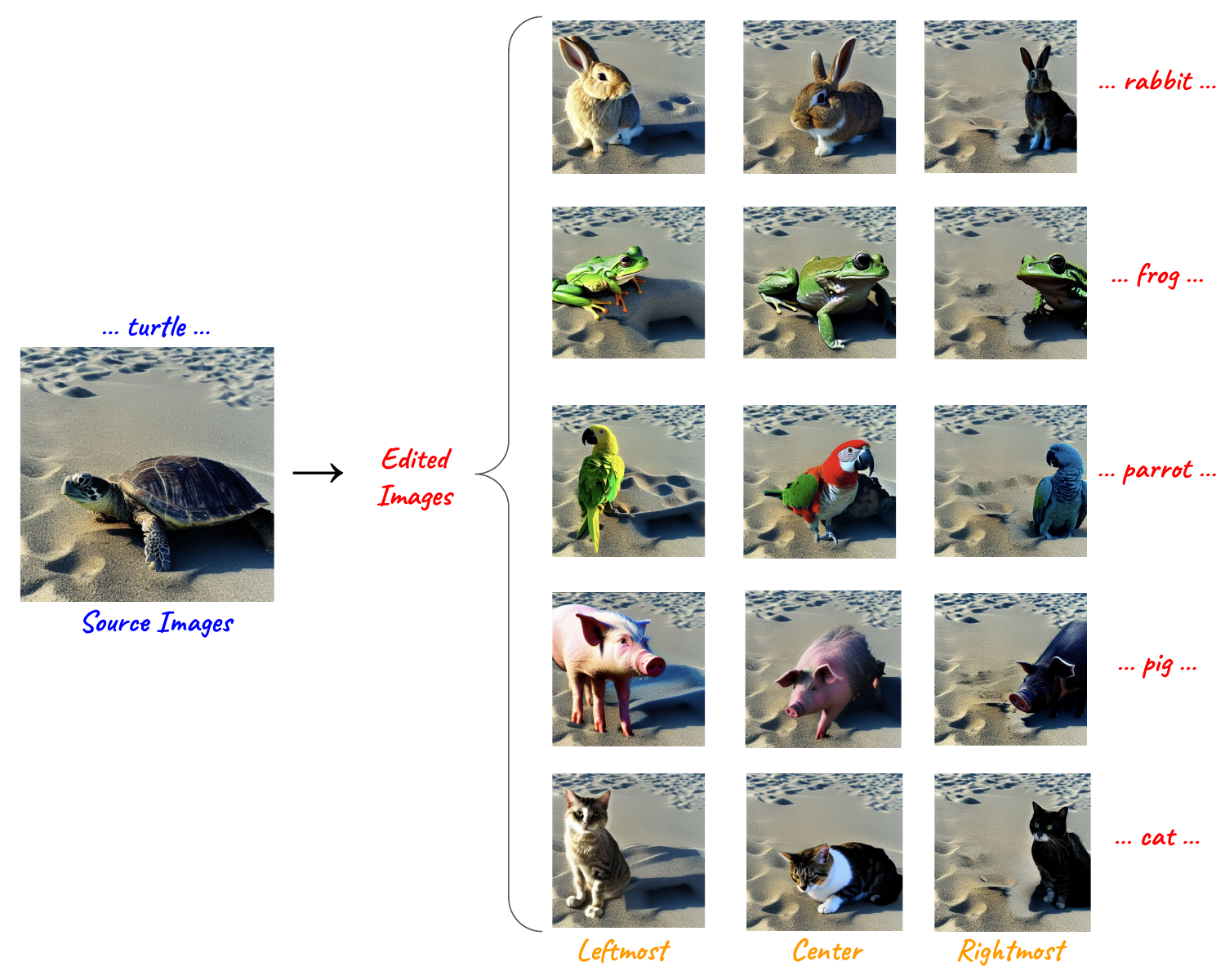}
    \caption{Visualization results of FlexEdit in achieving controllable object replacement with varying \textbf{position}.}
    \label{fig:control_position}
\end{figure}

\begin{figure}
    \centering
    \includegraphics[width=\textwidth]{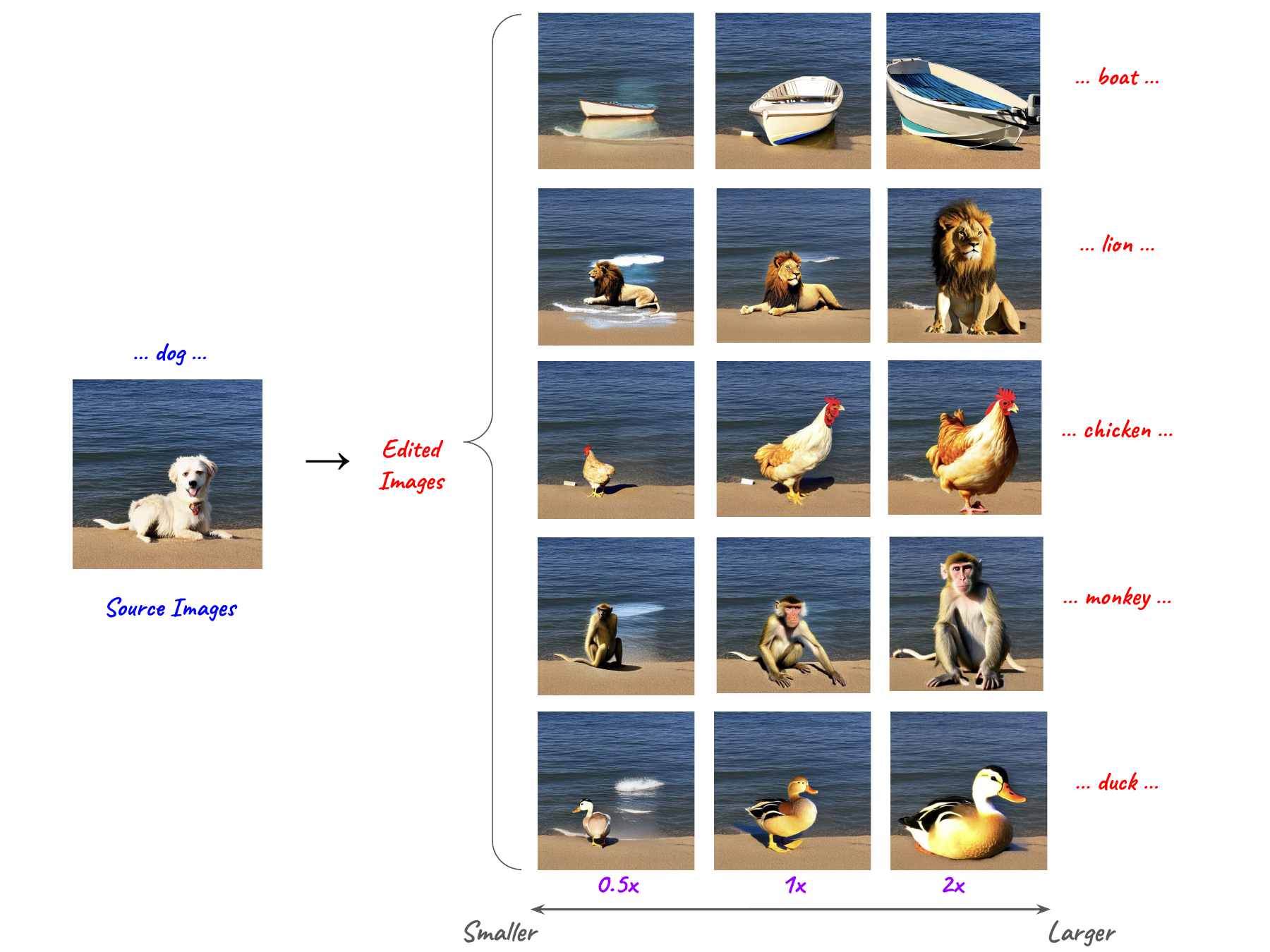}
    \caption{Visualization results of FlexEdit in achieving controllable object replacement with varying \textbf{size}.}
    \label{fig:control_size}
\end{figure}

In real-world scenarios, users might want to replace objects in a controllable manner. To this end, FlexEdit provides extensive control over objects' properties when the user can explicitly specify the relative position and size of a replaced object with respect to the source object. As shown in \cref{fig:control_position} and \cref{fig:control_size},
FlexEdit could handle diverse object replacement scenarios, ensuring that the properties of the replaced objects align with our specified constraints about the object's properties in terms of size or position while maintaining high fidelity to the source image.

\subsection{Mask-free Object Insertion}
\begin{figure}
    \centering
    \includegraphics[width=\textwidth]{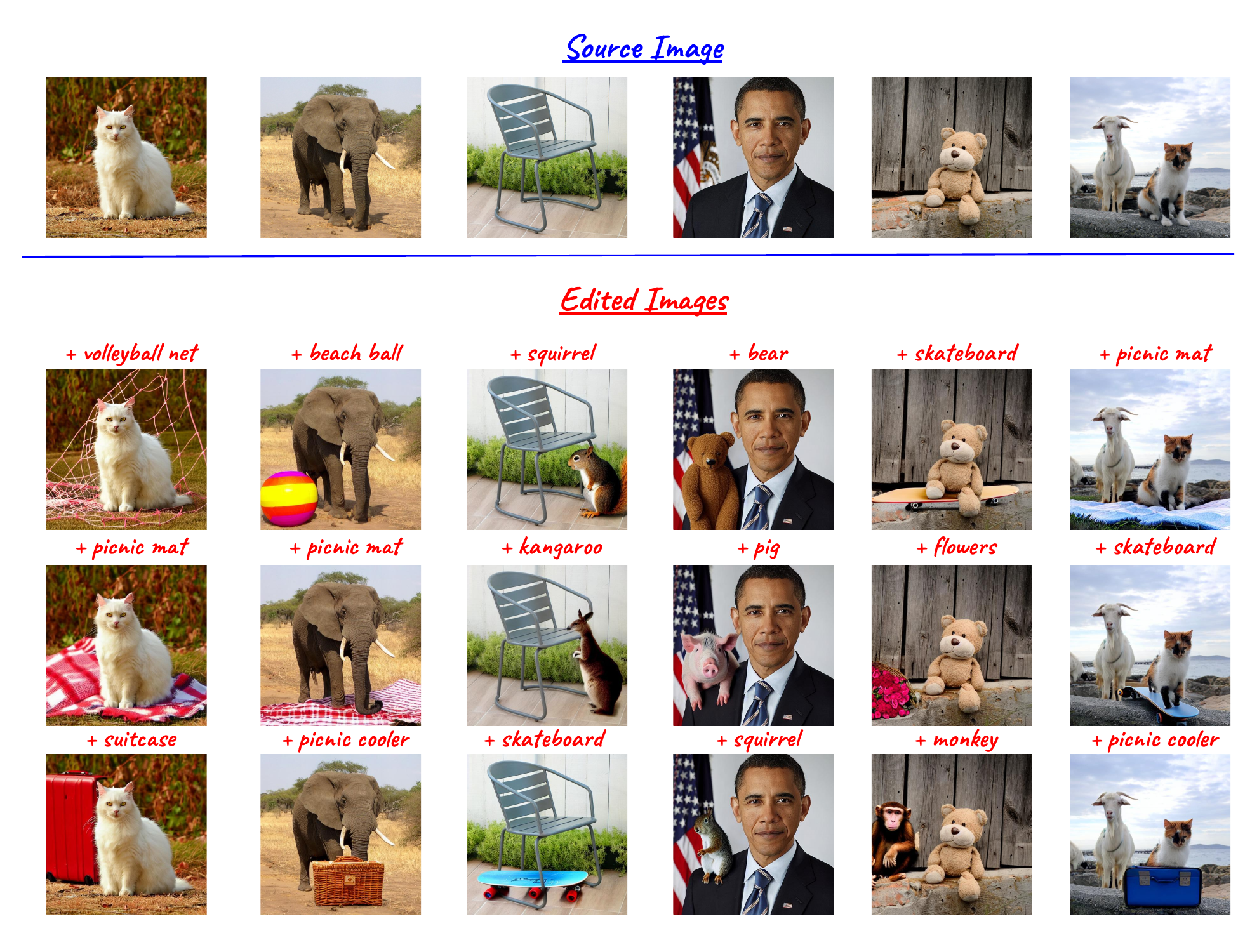}
    \caption{Visualization results of FlexEdit in achieving mask-free object addition.}
    \label{fig:add_without_mask}
\end{figure}

Another interesting application of FlexEdit is to perform editing by inserting new objects without explicitly requiring mask input from users. Our FlexEdit could add object via text guidance by specifying the target prompt as source prompt combined with additional objects that users might want to add. The method could automatically add the object to a reasonable location and with reasonable interactions with existing objects. As shown in \cref{fig:add_without_mask}, objects could be seamlessly inserted into the source image without affecting the original source image's content. For instance, when adding a ``picnic mat'' to a photo of a cat, FlexEdit automatically puts the mat on the ground and under the cat. It is layered well to seamlessly blend into the image context.


\clearpage  

%
%
\end{document}